%% file: main.tex
\documentclass[10pt,twocolumn,letterpaper]{article}

\usepackage[pagenumbers]{cvpr} 


\definecolor{cvprblue}{rgb}{0.21,0.49,0.74}
\usepackage[pagebackref,breaklinks,colorlinks,allcolors=cvprblue]{hyperref}
\usepackage{url}
\usepackage{booktabs}
\usepackage{array}
\usepackage{bm}
\usepackage{float}
\usepackage{graphicx}
\usepackage{wrapfig}
\usepackage{amssymb}
\usepackage{pifont}
\usepackage{multirow}
\usepackage{subcaption}
\usepackage{amssymb}
\usepackage{xcolor}

\def\paperID{15547} 
\def\confName{CVPR}
\def\confYear{2025}

\title{DreamRelation: Bridging Customization and Relation Generation}

\author{
\centerline{
Qingyu Shi\textsuperscript{\rm 1} \quad
Lu Qi\textsuperscript{\rm 3}$^{*}$ \quad
Jianzong Wu\textsuperscript{\rm 1} \quad
Jinbin Bai\textsuperscript{\rm 5} \quad
Jingbo Wang\textsuperscript{\rm 4} \quad
Yunhai Tong\textsuperscript{\rm 1} \quad
Xiangtai Li\textsuperscript{\rm 2,4}$^{*}$
} \\
\centerline{
\textsuperscript{\rm 1} PKU \quad
\textsuperscript{\rm 2} NTU \quad
\textsuperscript{\rm 3} Insta360 Research \quad
\textsuperscript{\rm 4} Shanghai AI Laboratory \quad
\textsuperscript{\rm 5} NUS \quad
}\\
\centerline{
Project page: \url{https://shi-qingyu.github.io/DreamRelation.github.io}
}
}

\begin{document}

\twocolumn[{
\renewcommand\twocolumn[1][]{#1}
\maketitle
\vspace{-8mm}
\centering
\includegraphics[width=0.98\textwidth]{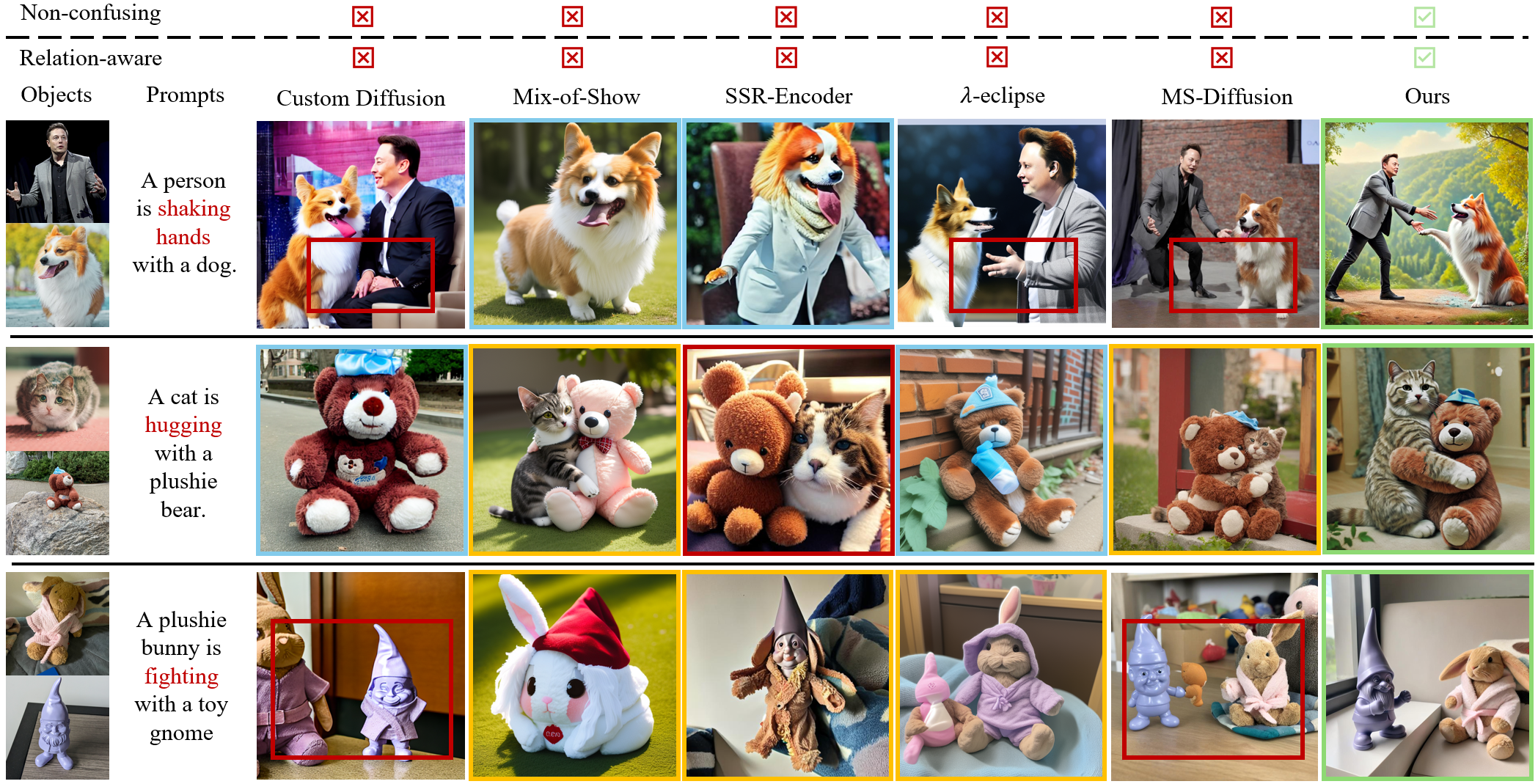}
\captionof{figure}{In our Relation-Aware Image Customization task, the generated images must accurately preserve the relationships between objects and maintain their identity. We highlight the limitations of previous approaches using three color codes: red indicates \textbf{failure to capture relationships}, blue marks \textbf{missing objects}, and orange represents \textbf{object confusion}. Each image is annotated to reflect its specific issue. Our results, highlighted by green boxes, demonstrate the advantages of our proposed method.}
\label{fig:teaser}
\vspace{0.7cm}
}]

\input{contents/0_abstract}    
\input{contents/1_introduction}
\input{contents/2_related_work}
\input{contents/3_method}
\input{contents/4_experiment}
\input{contents/5_conclusion}

{
    \small
    \bibliographystyle{ieeenat_fullname}
    \bibliography{main}
}

\input{contents/6_appendix}

\end{document}

%% file: contents/0_abstract.tex
\begin{abstract}
Customized image generation is essential for creating personalized content based on user prompts, allowing large-scale text-to-image diffusion models to more effectively meet individual needs.
However, existing models often neglect the relationships between customized objects in generated images.
In contrast, this work addresses this gap by focusing on relation-aware customized image generation, which seeks to preserve the identities from image prompts while maintaining the relationship specified in text prompts.
Specifically, we introduce DreamRelation, a framework that disentangles identity and relation learning using a carefully curated dataset.
Our training data consists of relation-specific images, independent object images containing identity information, and text prompts to guide relation generation.
Then, we propose two key modules to tackle the two main challenges—generating accurate and natural relationships, especially when significant pose adjustments are required, and avoiding object confusion in cases of overlap. 
First, we introduce a keypoint matching loss that effectively guides the model in adjusting object poses closely tied to their relationships.
Second, we incorporate local features of the image prompts to better distinguish between objects, preventing confusion in overlapping cases. 
Extensive results on our proposed benchmarks demonstrate the superiority of DreamRelation in generating precise relations while preserving object identities across a diverse set of objects and relationships.
The source code and trained models will be made available to the public.

\footnotetext[1]{* Corresponding authors: Lu Qi, Xiangtai Li.}
\vspace{-5mm}
\end{abstract}

%% file: contents/1_introduction.tex
\section{Introduction}
\label{sec:intro}

Driven by large-scale text-to-image diffusion models~\cite{stable_diffusion, imagen, sdxl, StableCascade}, customized image generation has recently made significant strides~\cite{dreambooth, customdiffusion, ELITE, ssr_encoder, ms-diffusion, anymaker, bai2024meissonic}. This task focuses on generating images that preserve the identity of objects from user-provided inputs, enabling the creation of personalized and meaningful content. It has shown value in numerous applications, including personalized artwork, branding, virtual fashion try-ons, social media content creation, augmented reality experiences, and more.

Despite the success of many methods for customizing single or multiple objects~\cite{cones2, ip-adapter, blip-diffusion, lambda_clipse, ms-diffusion, mix-of-show, clif, attndreambooth}, they often overlook the relationships between objects and the corresponding text prompts. For instance, when two user-provided objects are paired with a text prompt specifying a particular relationship, the generated output should not only preserve their identities but also accurately reflect the intended relationship, such as a `hug'. This introduces new challenges in what we refer to as relation-aware customized image generation, which focuses on preserving multiple identities while adhering to relationship prompts.

An intuitive solution to this issue is to adapt existing relation-aware generation methods~\cite{reversion, customizing_with_inverted_interaction, customizing_motion} to tuning-based or training-based customized methods for customizing objects while maintaining the relation. However, both approaches face challenges. Tuning-based methods struggle to preserve multiple identities, as they invert objects into specific tokens. While training-based schemes often fail to balance image and text prompts, frequently overlooking key textual elements and hindering the generation of relationships between objects. As illustrated in Fig.~\ref{fig:teaser}, previous methods fail to capture the relationships described in the text prompt and lose identity preservation.

We attribute this failure to two key factors: a lack of relevant data and an ineffective model design. Unlike data augmentation techniques such as flipping or rotation, commonly used to create paired training data in object customization methods~\cite{paint_by_example, anydoor}, our approach requires a triplet of images: two image prompts and one target image. The image prompts should contain similar objects but exhibit distinct poses compared to the target image. To collect these triplets, we propose a data engine to curate our fine-tuning set. We leverage an advanced text-to-image generation model~\cite{dalle3} to generate triplets where the same object pair is shared across the images. Through text prompt guidance, the image prompt provides strong identity information. This enables the decoupling of the relationship in the target image, which enhances the relation learning.

For the model design, we propose DreamRelation, which applies the Low-Rank Adaptation (LoRA)~\cite{lora} strategy to the text cross-attention layers of existing diffusion models to process user-provided text prompts. In DreamRelation, two key modules are introduced during training to enhance the relation generation in customized generation. 
First, we introduce a keypoint matching loss (KML) as additional supervision to explicitly encourage the model to manipulate object poses, since relationships between objects are closely tied to their poses. Importantly, the KML operates on the latent representation rather than the original image space, aligning with the default diffusion loss. 
Second, since relation-aware customization requires local features from image prompts, such as the ``hands'' features for generating ``shaking hands''—which are not captured by CLIP's coarse image-level features, we introduce dense features from CLIP. Through partitioning and pooling, we obtain local tokens that contain detail and local information. To further enhance the compatibility between dense features and image-level features, we employ a self-distillation method~\cite{clipself} to improve their alignment.

To more comprehensively evaluate relation-aware customized image generation, we constructed our RelationBench based on three established benchmarks \cite{dreambooth, customdiffusion}. Our methods demonstrate strong performance compared to existing approaches, achieving significant improvements in visual quality and quantitative results.

The contributions of this work are:
\begin{itemize}
    \item We explore a novel task called relation-aware customized generation, which aims to preserve multiple identities from image prompts while adhering to the relationships specified in text prompts. This task can enhance various user-driven applications by enabling more control.
    \item We introduce a data engine that uses the recent text-to-image generation model to generate triplet images where the same object is present with distinct poses. This well-curated dataset enables the model to focus on relation learning during fine-tuning, leveraging the identity information provided by the image prompts.
    \item Our proposed DreamRelation has two key modules, including keypoint matching loss and local token injection, to enhance the relation generation in customized generation. The extensive experiments on three benchmarks demonstrate the effectiveness of our method.
\end{itemize}

%% file: contents/2_related_work.tex
\section{Related Work}
\label{sec:related_work}

\noindent
\textbf{Diffusion-Based Text-to-Image Generation.} Diffusion-based text-to-image models~\cite{stable_diffusion, imagen} generate high-quality images based on user-provided text prompts. These works encode the text prompt through the text encoder~\cite{clip, t5} and inject the text embedding into the U-Net's cross-attention layers. Some methods~\cite{sdxl, StableCascade, ultrapixel, wu2024motionbooth} upscale diffusion models and incorporate additional conditions as priors to generate high-resolution images. Meanwhile, several methods~\cite{dalle3, llm_grounded_diffusion} introduce stronger text encoders or large language models (LLMs) to enhance the text comprehension capabilities of diffusion models. Recent works \cite{DiT, pixart} have replaced the U-Net denoiser with the Transformers~\cite{attention}. 
However, these models primarily focus on text conditions and struggle to handle other forms of guidance, such as user-provided images. In contrast, our work facilitates customized generation while following relation conditions given by text inputs.

\noindent
\textbf{Diffusion-Based Customized Generation.} 
Customized image generation aims to produce new images based on user-provided objects. Tuning-based methods accomplish this by fine-tuning specific parameters of diffusion models to learn new objects. 
Some of these methods~\cite{textual_inversion, p+, cones2} employ text embeddings to represent the object identity.
On the other hand, DreamBooth~\cite{dreambooth} fine-tunes the entire U-Net and introduces a prior preservation loss to mitigate language drift.
Additionally, several approaches~\cite{customdiffusion, mix-of-show, clif, attndreambooth} integrate text embeddings and U-Net parameters for image customization. These methods fine-tune both components during object learning, achieving impressive results.
Considering the substantial cost of fine-tuning for commercial uses, training-based methods~\cite{blip-diffusion, ELITE, anydoor, lambda_clipse, ssr_encoder} have been proposed. This approach primarily utilizes an encoder to extract the object's identity and inject it into the U-Net. For instance, MS-Diffusion~\cite{ms-diffusion} integrates grounding tokens with a Resampler to maintain detailed identity and employs layout guidance to explicitly locate the objects. 
Despite their efficiency during inference, training-based methods often face challenges in balancing identity preservation and text control. A common issue is that the relationships described in text prompts are frequently overlooked. To address this limitation, we introduce DreamRelation, a framework designed to bridge the customization and relation generation.

\noindent
\textbf{Relation-Aware Text-to-Image Generation.} Inspired by tuning-based customization methods~\cite{textual_inversion}, some recent works~\cite{reversion, dreamvideo, customizing_with_inverted_interaction} represent a ``neglected word'' by learnable parameters. 
These methods fine-tune part of the parameters on the content co-existing images. For instance, Reversion~\cite{reversion} fine-tunes the text embedding on a set of relation co-existing images and introduces a relation-steering contrastive loss. Reversion gains a better alignment between relational words and generated images by injecting new text embedding into prompts. While effective with prepositions and adjectives, Reversion struggles with relationships that involve significant overlaps, and cannot customize user-provided objects. Moreover, this method still struggles to generate vivid relationships while maintaining the fidelity of multiple customized objects, likely due to the inherent disregard for text embeddings in customization methods.
Our work bridges the gap in the predicate relation-following capability of training-based customization methods. 
This enables efficient inference that naturally and accurately generates relationships in the text prompts.

%% file: contents/3_method.tex
\section{Bridging Customization and Relation Generation}
\label{sec:method}

\begin{figure*}[t]
  \centering
\includegraphics[width=1.0\textwidth]{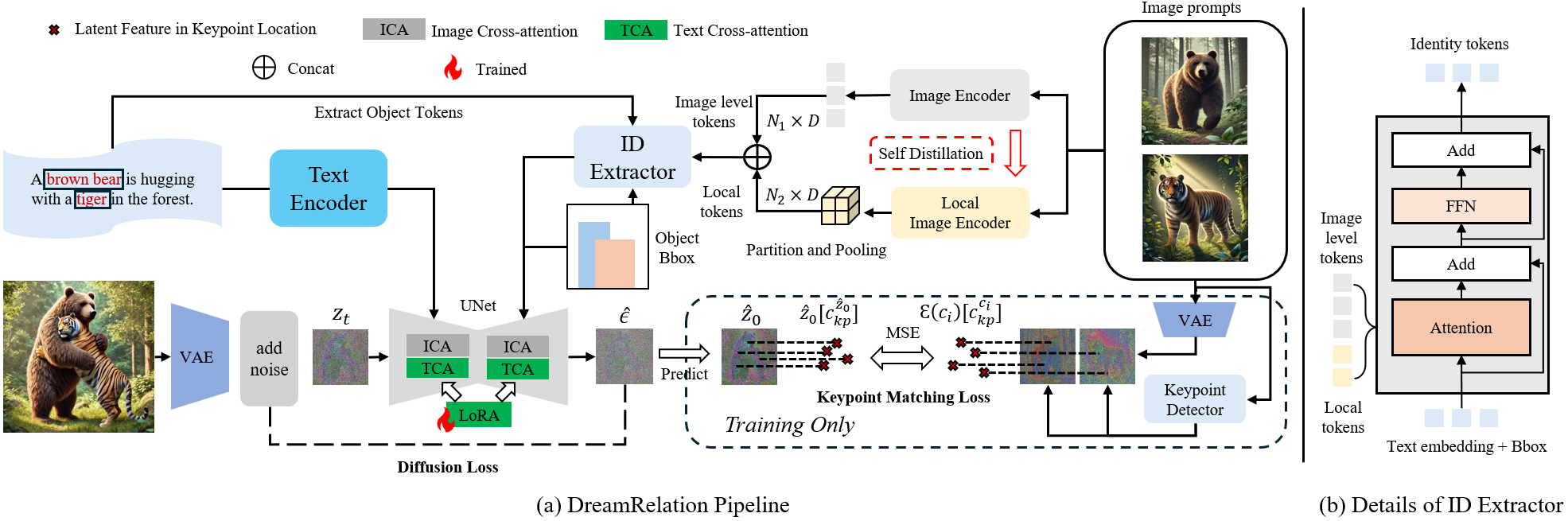}
\vspace{-16pt}
  \caption{The overview of DreamRelation. DreamRelation utilizes the off-the-shelf identity extractor to decouple the relation and identity information in relation-specific images. After getting the U-Net output $\hat{\epsilon}$, we predict $\hat{z}_0$ and calculate the keypoint matching loss. The part in the dotted box is only for training.}
  \label{fig:pipeline}
  \vspace{-6pt}
\end{figure*}

\subsection{Problem definition}
Different from conventional customization tasks, we explore a new setting that focuses on image generation by both image prompts $c_i \in \mathcal{R}^{N\times 3\times H\times W}$ and text prompts $c_t$.
We call this setting relation-aware customized image generation due to the requirements that the generated image $\hat{x}$ should strictly preserve each identity and keep the relationship among those identities provided by $c_i$ and $c_t$.

We define this task as:
\begin{equation}
\hat{x} = \Phi_{\theta}(c_i, c_t)
\end{equation}

where $\Phi_{\theta}$ is the network parameterized by $\theta$. For brevity clarification, we set $N=2$ due to the basic triplet element defined in the DreamRelation.

\subsection{Discussion and Motivation}
While previous works~\cite{customdiffusion, mix-of-show, ssr_encoder} have explored the customization of multiple objects under text control, they lack the ability to effectively manage the relationships between these objects.
Specifically, the state-of-the-art multi-object customization model~\cite{ms-diffusion} utilizes bounding boxes to enhance multi-object generation. Given the image prompt \( c_i \) and the text prompt \( c_t \), it leverages CLIP~\cite{clip} and the Resampler~\cite{ms-diffusion} to extract text tokens \( tok_\text{text} \) and image tokens \( tok_\text{image} \), respectively. These tokens are then injected into the U-Net \( \bm{\epsilon_\theta} \) through parallel cross-attention layers:
\begin{equation}
    h =\text{Softmax}(\frac{QK_i}{\sqrt{d}} + \mathcal{M})V_i + \text{Softmax}(\frac{QK_t}{\sqrt{d}})V_t
\end{equation}
where \( Q \) represents the query, while \( K_i, V_i \) and \( K_t, V_t \) denote the keys and values obtained from \( tok_\text{image} \) and \( tok_\text{text} \), respectively. The mask \( \mathcal{M} \), generated from the bounding boxes, assists the model in localizing objects during multi-object customization. However, it encounters difficulties when bounding boxes overlap, frequently resulting in object confusion, as shown in Fig.~\ref{fig:teaser}. Consequently, relation-aware generation in customized models remains a largely unexplored area of research.

On the other hand, we revisit the relation-aware generation task. Existing relation inversion methods~\cite{reversion, customizing_with_inverted_interaction} focus on inverting a relationship into a text embedding \( R^* \) within pre-trained text-to-image diffusion models. Given a set of images \(\{x_k\}_{k=1}^{n}\) that share the same relationship, these methods use a denoising loss \(\mathcal{L}_{RI}\) to fine-tune \( R^* \), ensuring alignment with the specific relationship:
\begin{equation}
    \mathcal{L}_{RI} = \mathbb{E}_{z_t, t, \epsilon, c}[\Vert \epsilon_t - \bm{\epsilon_{\theta}}(z_t, t, c_t) \Vert_{2}^{2}]
\end{equation}

Directly applying relation-aware generation methods to customized models leads to suboptimal performance, as illustrated in Fig.~\ref{fig:abl_data}. We attribute this to two main reasons. First, the balance between the control of \( c_i \) and \( c_t \) is not addressed in relation-aware generation tasks. During inference, the model tends to overlook the target relationship \( R^* \) due to the influence of \( c_i \), resulting in incorrect relationship generation. Second, even if the shared relationship exists across \(\{x_k\}_{k=1}^{n}\), the disentangled object identity information complicates the learning process. Fine-tuning solely on \(\{x_k\}_{k=1}^{n}\) makes it challenging for the model to effectively capture the relationship. As shown in the second row of Fig.~\ref{fig:abl_data}, where ReVersion~\cite{reversion} struggles to generate the "hugging" relationship even in the absence of \( c_i \). In the following subsections, we will introduce our data collection method and DreamRelation to bridge customization and relation generation. 

\subsection{Data Collection}
To address the balance between $c_i$ and $c_t$ while facilitating relation learning from \(\{x_k\}_{k=1}^{n}\), we first design a data engine for collecting high-quality tuning data. 
The ideal tuning data should be in the form of $\mathcal{D} = (x_k, c_i, c_t)$. The $c_i$ contain the object in $x_k$ while $c_t$ describes the relationship between objects in $x_k$. The identity information in \( c_i \) helps decouple the relationship within \( x_k \), enhancing relation learning while maintaining a balanced integration of \( c_i \) and \( c_t \).
Unfortunately, directly cropping the objects from $x_k$ to obtain $c_i$ results in a copy-and-paste effect~\cite{anydoor}, even when using data augmentation techniques like flipping and rotation. Therefore, we introduce our data engine below.

\begin{figure}[t]
  \centering
  \includegraphics[width=0.50\textwidth]{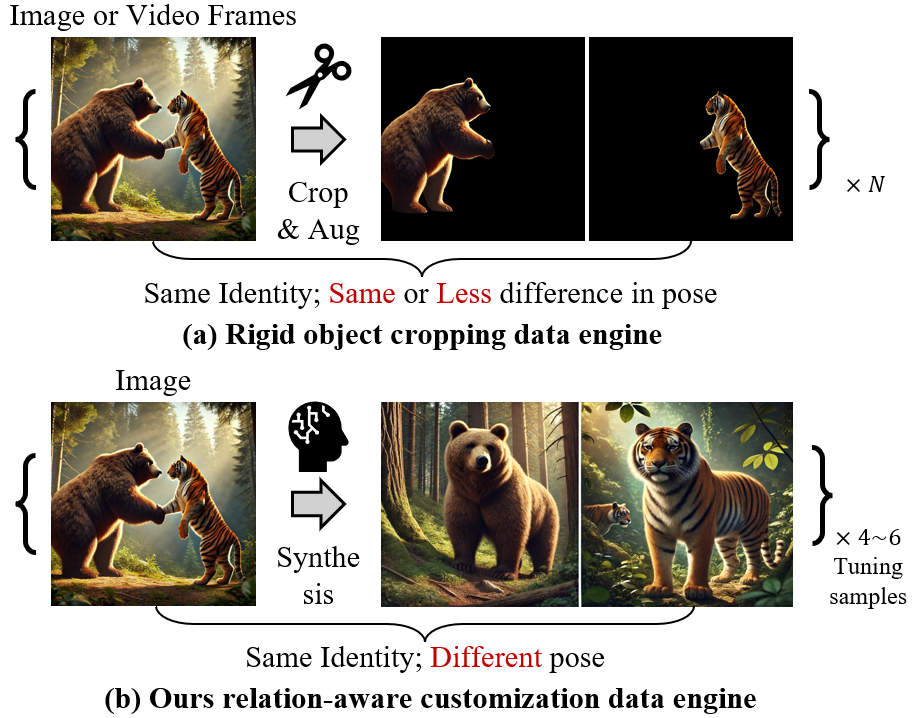}
  \caption{The rigid object cropping data engine, which results in minimal changes to object pose, the cropped image prompts also contain relation information, leading to a copy-and-paste effect, where the text prompt is neglected. Our relation-aware data engine, on the other hand, focuses on relation learning by decoupling identity information from images.}
  \vspace{-10pt}
  \label{fig:data_engine}
\end{figure}

\noindent
\textbf{Relation-aware data engine.}
Fig.~\ref{fig:data_engine} illustrates the difference of our data engine. Both of \( x_k \) and \( c_i \) are generated by the recent text-to-image generation model, rather than through cropping and augmentation. Specifically, inspired by the performance of DALL-E 3 and its multi-turn dialog capability. We use DALL-E 3 to generate $\mathcal{D} = (x, c_i, c_t)$. Where $x_k$ and $c_i$ share the same object identity by leveraging the prompt ``The photo of the same.''
We observe that prompting in this manner allows DALL-E 3 to remember and preserve identity in common categories such as "tiger", which is enough for relation learning. 
Next, we employ X-Pose~\cite{unipose}, SAM~\cite{kirillov2023sam}, and LLaVA~\cite{llava} to annotate $x$, $c_i$ with keypoints, masks, and captions for samples.

\subsection{DreamRelation}

\subsubsection{Relation Learning}
\begin{figure}[t]
  \centering
  \includegraphics[width=0.45\textwidth]{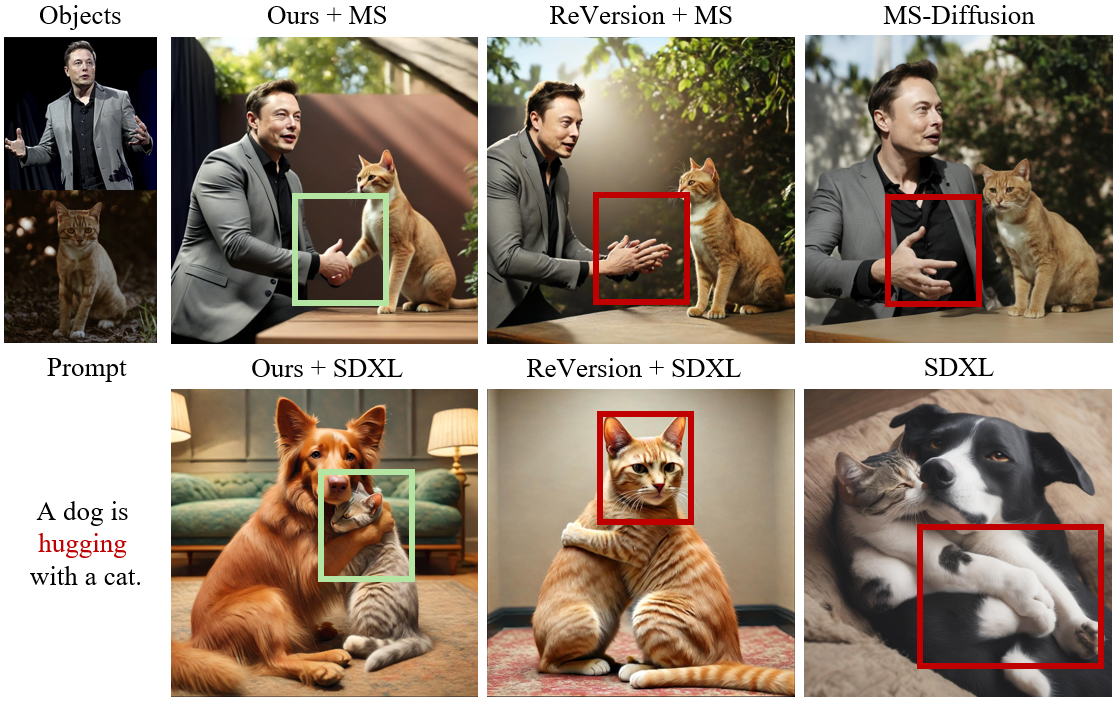}
  \vspace{-5pt}
  \caption{Comparing our method with ReVersion across different base models, our approach demonstrates superior performance in the relation-aware generation task.}
  \vspace{-5pt}
  \label{fig:abl_data}
\end{figure}

To balance \( c_i \) and \( c_t \) while enhancing relation learning, we decouple the identity information and relationships in \( x_k \). Specifically, we leverage an off-the-shelf ID extractor from the state-of-the-art customized model~\cite{ms-diffusion}. The identity information in \( c_i \) enables the model to focus on the relationship information in \( x_k \) and \( c_t \) during relation learning. To achieve this, we employ parallel cross-attention layers to process \( c_i \) and \( c_t \), which can be formulated as:
\begin{equation}
    h = \gamma \cdot \text{Softmax}(\frac{QK_i}{\sqrt{d}} + \mathcal{M})V_i + \text{Softmax}(\frac{QK_t}{\sqrt{d}})V_t
\end{equation}
where $Q = W_q h$, $K_i = W_{k_i} c_i$, $V_i = W_{v_i} c_i$, $K_t = W_{k_t} c_t$, $V_t = W_{v_t} c_t$, and $\gamma$ is a hyperparameter for scaling control from $c_i$. For clarity, we omit the final linear layer $W_{out}$. 
The relation information mainly depends on text prompts. Therefore, we inject LoRA layers in all of the $W_q$, $W_{k_t}$, $W_{v_t}$, and $W_{out}$ to encourage the model to pay more attention to the relationship in $c_t$.
We freeze all other parameters during fine-tuning. 
As shown in Fig.~\ref{fig:abl_data}, benefiting from the decoupled identity information in $c_i$, our method effectively captures the relationship in $x_k$ and $c_t$, accurately generating the corresponding images. Notably, after fine-tuning, our method can also be directly integrated into SDXL~\cite{sdxl}, effectively enabling the model to address the relation-aware generation task. The extensive results are presented in the Supplementary Material (SM).

\begin{table}[t!]
   \centering
   \scalebox{0.56}{
   \setlength{\tabcolsep}{2mm}
   \begin{tabular}{c c c c c}
      \toprule[0.15em]
        Methods & Single-Object Cus & Multi-Object Cus & Non-Confusing & Relation-Aware \\
        \hline
        Textual Inversion~\cite{textual_inversion} & \textcolor{blue}{\checkmark} & \textcolor{red}{\ding{55}} & \textcolor{red}{\ding{55}} & \textcolor{red}{\ding{55}} \\
        DreamBooth~\cite{dreambooth} & \textcolor{blue}{\checkmark} & \textcolor{red}{\ding{55}} & \textcolor{red}{\ding{55}} & \textcolor{red}{\ding{55}} \\
        Custom Diffusion~\cite{customdiffusion} & \textcolor{blue}{\checkmark} & \textcolor{blue}{\checkmark} & \textcolor{red}{\ding{55}} & \textcolor{red}{\ding{55}} \\
        Mix-of-Show~\cite{mix-of-show} & \textcolor{blue}{\checkmark} & \textcolor{blue}{\checkmark} & \textcolor{red}{\ding{55}} & \textcolor{red}{\ding{55}} \\
        CLIF~\cite{clif} & \textcolor{blue}{\checkmark} & \textcolor{blue}{\checkmark} & \textcolor{blue}{\checkmark} & \textcolor{red}{\ding{55}} \\
        MultiBooth~\cite{multibooth} & \textcolor{blue}{\checkmark} & \textcolor{blue}{\checkmark} & \textcolor{red}{\ding{55}} & \textcolor{red}{\ding{55}} \\
        \hline
        BLIP-Diffusion~\cite{blip-diffusion} & \textcolor{blue}{\checkmark} & \textcolor{red}{\ding{55}} & \textcolor{red}{\ding{55}} & \textcolor{red}{\ding{55}} \\
        ELITE~\cite{ELITE} & \textcolor{blue}{\checkmark} & \textcolor{red}{\ding{55}} & \textcolor{red}{\ding{55}} & \textcolor{red}{\ding{55}} \\
        AnyDoor~\cite{anydoor} & \textcolor{blue}{\checkmark} & \textcolor{red}{\ding{55}} & \textcolor{red}{\ding{55}} & \textcolor{red}{\ding{55}} \\
        SSR-Encoder~\cite{ssr_encoder} & \textcolor{blue}{\checkmark} & \textcolor{blue}{\checkmark} & \textcolor{red}{\ding{55}} & \textcolor{red}{\ding{55}} \\
        $\lambda$-eclipse~\cite{lambda_clipse} & \textcolor{blue}{\checkmark} & \textcolor{blue}{\checkmark} & \textcolor{red}{\ding{55}} & \textcolor{red}{\ding{55}} \\
        MS-Diffusion~\cite{ms-diffusion} & \textcolor{blue}{\checkmark} & \textcolor{blue}{\checkmark} & \textcolor{red}{\ding{55}} & \textcolor{red}{\ding{55}} \\
        \hline
        ReVersion~\cite{reversion} & \textcolor{red}{\ding{55}} & \textcolor{red}{\ding{55}} & \textcolor{blue}{\checkmark} & \textcolor{blue}{\checkmark}  \\
        ADI~\cite{adi} & \textcolor{red}{\ding{55}} & \textcolor{red}{\ding{55}} & \textcolor{red}{\ding{55}} & \textcolor{blue}{\checkmark}  \\
        \hline
        \textbf{DreamRelation} & \textcolor{blue}{\checkmark} & \textcolor{blue}{\checkmark} & \textcolor{blue}{\checkmark} & \textcolor{blue}{\checkmark} \\
      \bottomrule[0.10em]
   \end{tabular}}
   \vspace{-5pt}
   \caption{Setting Comparison for Different Models. We include several representative methods here. Our proposed DreamRelation can meet the demand of the relation-aware customization task.}
   \vspace{-5pt}
   \label{tab:model_scope}
\end{table}

\subsubsection{Keypoint Matching Loss}

Most relationships have specific requirements for the pose, such as "hugging" requiring the arms to cross over the object's body, and "riding a bicycle" requiring the feet of the object to be on the pedal plate. Therefore, it's important to manipulate the pose of objects in the right status accurately. Intuitively, we introduce explicit supervision in the diffusion latent space during the tuning process, named Keypoint Matching Loss (KML).

Specifically, considering our task only involves common objects instead of humans, we employ X-Pose~\cite{unipose} as a keypoint detector because it can detect any keypoints in complex real-world scenarios. We detect 17 keypoints for each object in $x_k$ and $c_i$, and denote the keypoint coordinates as $c^{x_k}_{kp}, c^{c_i}_{kp} \in \mathcal{R}^{17 \times 2}$ respectively. 
To encourage the model to generate accurate poses in $\mathcal{D}(\hat{z}_0)$, where $\mathcal{D}$ is the VAE decoder. We use the U-Net's output $\hat{\epsilon}$ to predict $z_0$ during fine-tuning:
\begin{equation}
    \hat{z}_0 = \frac{z_t - \sqrt{1 - \bar{\alpha}_t} \hat{\epsilon}}{\sqrt{\bar{\alpha}_t}}
\end{equation}

We use the VAE encoder $\mathcal{E}$ to obtain the latent representation of image prompts $\mathcal{E}(c_i)$. Then we calculate the MSE loss on the corresponding keypoint locations between $\mathcal{E}(c_i)$ and $\hat{z}_0$:
\begin{align}
    \mathcal{L}_{KML} &= \frac{1}{n_{kp}} \mathbb{E}_{\bm{z_t},c_i} \left \| \mathcal{E}(c_i)[c^{c_i}_{kp}] - \hat{z}_0[c^{x_k}_{kp}]  \right \|^2_2 \\
    \mathcal{L} &= \mathcal{L}_{denoise} + \lambda \cdot \mathcal{L}_{KML}
\end{align}
where $n_{kp}$ denote the number of keypoint, $c^{c_i}_{kp}$ and $c^{x_k}_{kp}$ are keypoint's coordinates in $c_i$ and $x_k$ respectively. We scale the coordinates to fit the size of $\mathcal{E}(c_i),\, \hat{z}_0$ $\lambda$ controls for the relative weight of KML. 
The part inside the dotted box in Fig.~\ref{fig:pipeline} illustrates the model fine-tuning with the KML. 
We find KML is effective in encouraging the model to manipulate the pose for relation-aware generation. 

\begin{figure*}[t]
  \centering
  \includegraphics[width=0.80\textwidth]{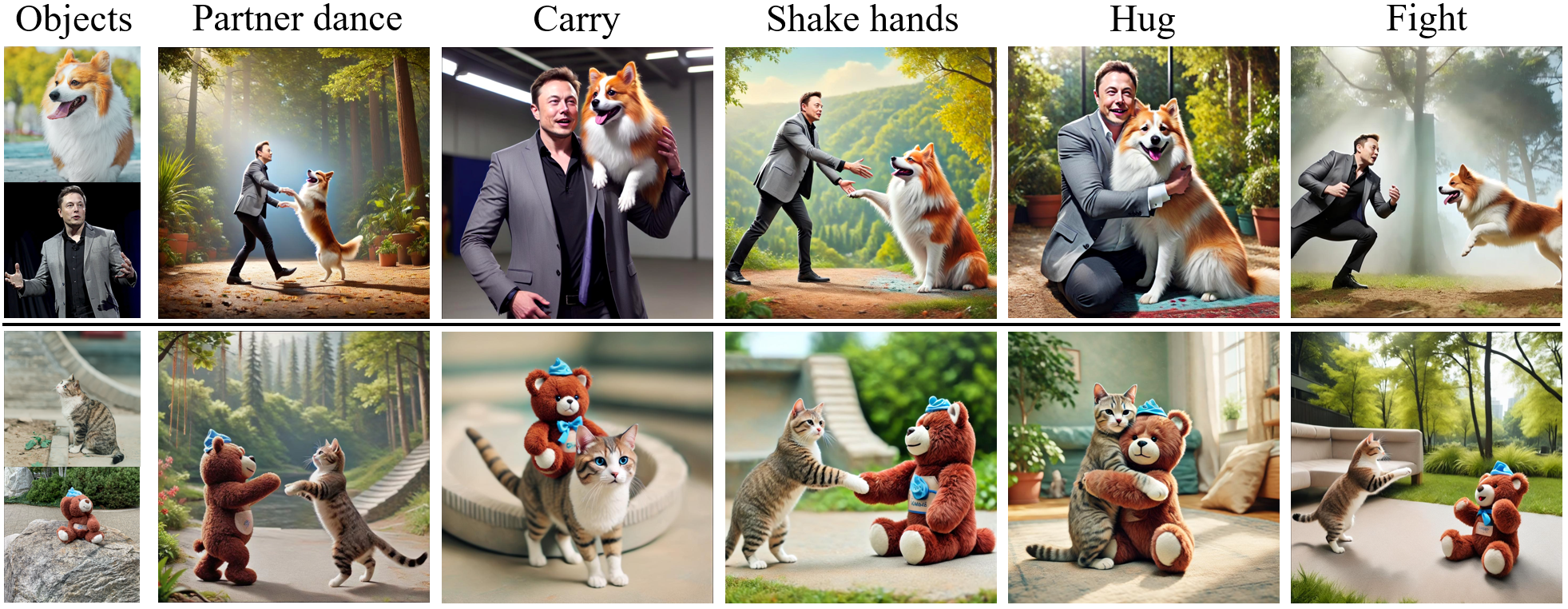}
  \caption{Additional results demonstrate the effectiveness of relation-aware generation. DreamRelation adapts the same pair of objects to different relationships in a natural and accurate manner.}
  \vspace{-10pt}
  \label{fig:main_results_multi}
\end{figure*}

\subsubsection{Local Token Injection}
When generating relationships, local and detailed features are essential—for instance, the feature of arms to accurately construct relationships like ``carrying''. However, the coarse image-level features extracted by the CLIP Image Encoder fail to provide such fine-grained details, leading to confusion between objects in Fig.~\ref{fig:abl}. To address this limitation, we introduce dense features from the CLIP Image Encoder~\cite{dense_clip}. Specifically, we modify the last layer of the CLIP Image Encoder to obtain dense features: \( h_{\text{dense}} = \text{ModifiedAttention}(h'_{\text{clip}}) \), where \( h'_{\text{clip}} \) represents the hidden states of the second-to-last layer.
\begin{align}
    h_{\text{tmp}} &= \text{Proj}_{v}(\text{norm}(h'_{\text{clip}})), \\
    h_{\text{tmp}} &= h'_{\text{clip}} + \text{Proj}_{out}(h_{\text{tmp}}), \\
    h_{\text{dense}} &= h_{\text{tmp}} + \text{FFN}(h_{\text{tmp}})
\end{align}

To further enhance the compatibility between dense features and image-level features, we employ self-distillation on the CLIP Image Encoder~\cite{clipself} to derive our Local Image Encoder. Specifically, we align local regions of the dense features with their corresponding image-level features using cosine similarity.

During inference, we extract local tokens \( tok_{\text{local}} \) from the dense feature \( h_{\text{dense}} \) through partitioning and pooling. These local tokens are then concatenated with the image-level tokens \( tok_{\text{image}} \) and passed to the ID extractor. The ID extractor, which is a transformer-based architecture as illustrated in Fig.~\ref{fig:pipeline} (b), can be written as:
\begin{equation}
    q = q + \text{Attention}(\text{concat}[q,\, tok_\text{image},\, tok_\text{local}])
\end{equation}

Regarding the compatibility between CLIP dense features and the original image-level features, experimental results demonstrate that our Local Image Encoder outperforms other dense representations~\cite{dino}. For a detailed discussion, we refer readers to the Supplementary Material. Additionally, the injection of local tokens enhances identity preservation, as evidenced by improvements in evaluation metrics and visual results, shown in Fig.~\ref{fig:abl} and Tab.~\ref{tab:main_ablation}.

%% file: contents/4_experiment.tex
\section{Experiments}
\label{sec:exp}

\begin{figure*}[t]
  \centering
  \includegraphics[width=1.0\textwidth]{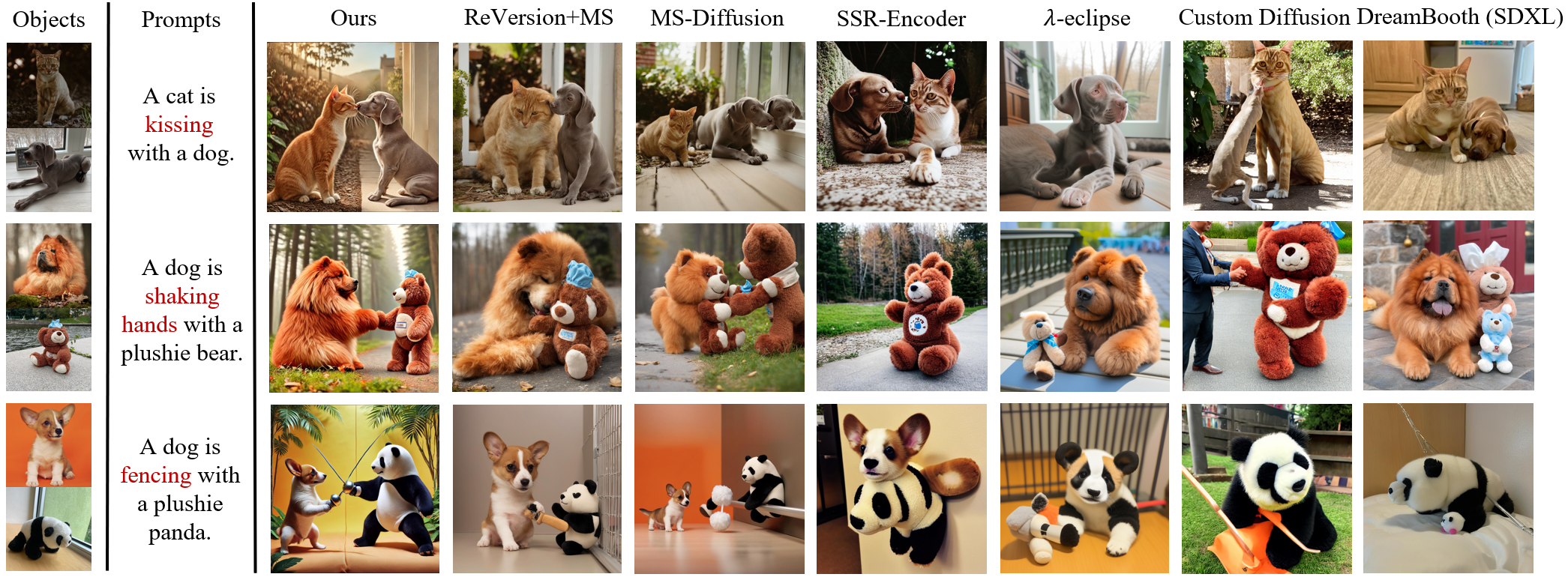}
  \vspace{-15pt}
  \caption{We compare multi-object performance with training-based and tuning-based methods. Additionally, we integrate ReVersion~\cite{reversion} into MS-Diffusion, where the learnable text token is injected into the text prompt during MS-Diffusion's inference, resulting in ReVersion+MS. Our method shows a clear advantage in relation-aware generation and avoiding object confusion in overlapping scenarios.}
  \vspace{-5pt}
  \label{fig:multi_comparison}
\end{figure*}

\begin{table*}[t]
    \centering
    \scalebox{1.0}
    {\input{tables/quantitative_comparison}}
    \vspace{-5pt}
    \caption{Quantitative comparison on RelationBench, \textbf{Bold} and \underline{underline} represent the highest and second-highest metrics.}
    \vspace{-10pt}
    \label{tab:comparison}
\end{table*}

\begin{figure}[t]
    \centering
    \includegraphics[width=0.45\textwidth]{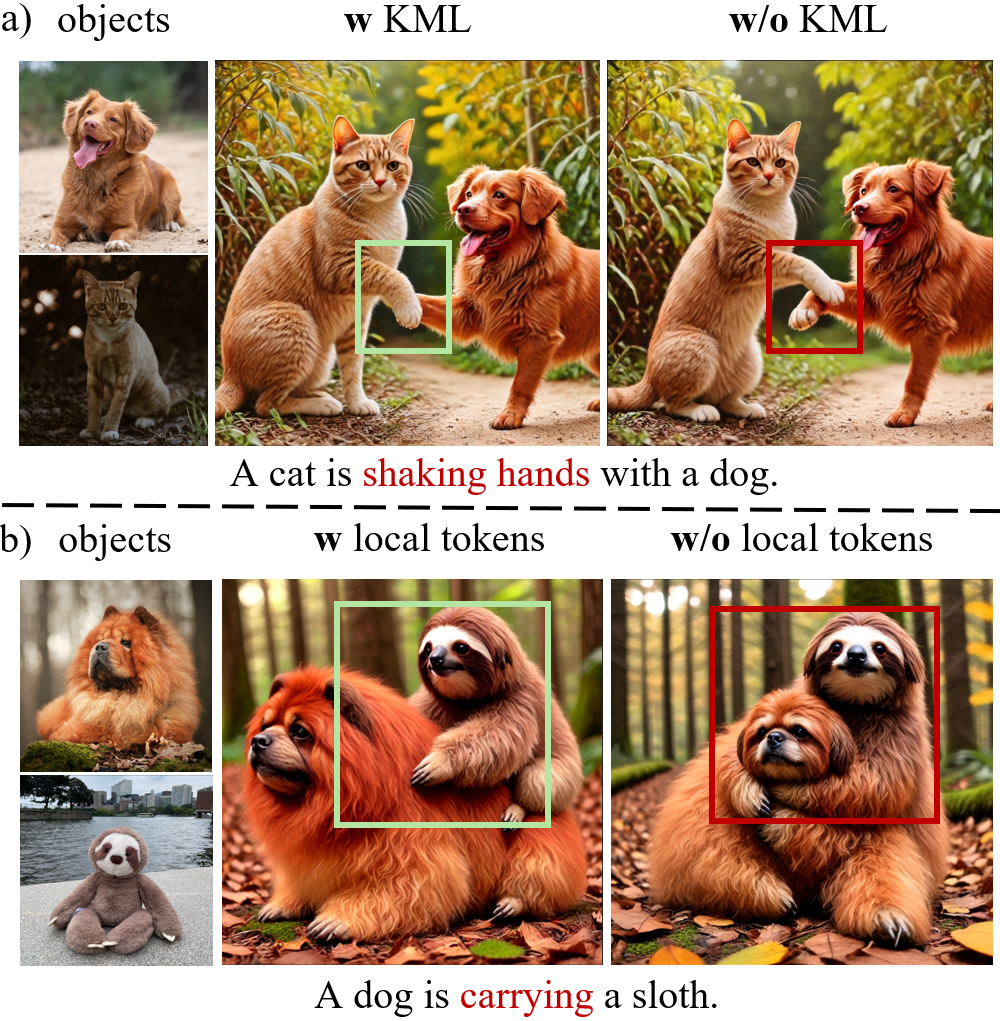}
    \caption{Ablation study of keypoint matching loss (KML) and local tokens: The images are generated using the same random seed. We use green and red boxes to highlight the main differences.}
    \vspace{-10pt}
  \label{fig:abl}
\end{figure}

\begin{figure}[t]
    \centering
    \includegraphics[width=0.45\textwidth]{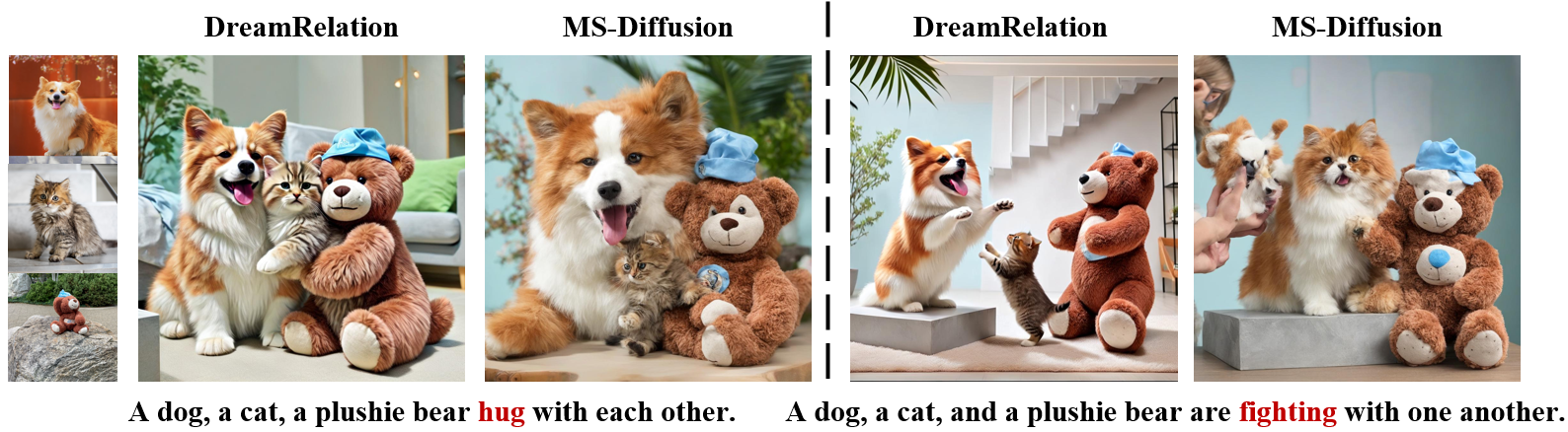}
    \caption{DreamRelation for three objects. Compared to the baseline model, DreamRelation shows a clear advantage.}
    \vspace{-10pt}
  \label{fig:3objects}
\end{figure}

\begin{table}[t]
    \centering
    \scalebox{0.79}
    {\input{tables/ablation_main}}
    \vspace{-5pt}
    \caption{Ablation study on our proposed components}
    \label{tab:main_ablation}
    \vspace{-15pt}
\end{table}

\noindent
\textbf{Implementation Details.} 
We generate a tuning set for relation learning with 4-6 samples per relationship. We incorporate LoRA layers into all text cross-attention layers of the U-Net. We set the LoRA rank to $r=4$, the parallel cross-attention scaling factor to $\gamma=0.6$, and the keypoint matching loss weight to $\lambda=1e-3$. The model is fine-tuned for 500 steps, using 2 A100 GPUs, with a total batch size of 8, completing the process in 10 minutes. We use the Adam optimizer~\cite{adam} with a learning rate of $1e-4$ and no weight decay, resulting in a total of 3.1M trainable parameters. Note that our learnable parameters are only in the text cross-attention layers, making it compatible with any SDXL-based models. Considering the strong identity preservation capabilities of MS-Diffusion, we implement our DreamRelation on MS-Diffusion. During fine-tuning and inference, we concatenate local tokens with image tokens. We put more self-distillation details of the Local Image Encoder in the Supplementary Material.

\noindent
\textbf{Evaluation.} 
To evaluate relation-aware customized image generation, we propose RelationBench, consisting of 44 objects from DreamBench~\cite{dreambooth} and CustomConcept101~\cite{customdiffusion}, along with 25 relationships. The object categories include pets, plushies, toys, people, and cartoons. Using GPT-4, we generate 100 cases for single-object and multi-object evaluations. Following previous works~\cite{dreambooth, ms-diffusion}, we evaluate our method on three metrics: (1) Identity Preservation, which assesses the alignment between the generated images and the image prompts, using the CLIP image score and DINO score that calculate the cosine similarity between the class embeddings of images, referred to as CLIP-I and DINO, respectively; (2) Text Alignment, which evaluates how well the generated images align with the text prompts, using the CLIP image-text score, denoted as CLIP-T; and (3) Relation Alignment. We observed that nouns in the prompts can inflate CLIP-T scores, making them less accurate for evaluating relation generation. To address this, we extract the relation word from the text prompt using spaCy and calculate the CLIP image-text score, denoted as CLIP-R.

\subsection{Main Results}
We evaluate our method through both quantitative and qualitative results in single- and multi-object cases.
In single-object cases, we primarily focus on the alignment of the object's pose with the relationship in the text prompt and the model's ability to preserve identity.
In multi-object cases, in addition to relation generation and identity preservation, we also need to ensure no confusion between the multiple objects.

\noindent
\textbf{Quantitative results.} We compare our method with baseline models across three benchmarks: RelationBench, DreamBench, and multi-object cases in CustomConcept101. As shown in Tab.~\ref{tab:comparison}, our method demonstrates a clear advantage on RelationBench for single-object cases, particularly in the CLIP-T and CLIP-R metrics. However, its performance is slightly lower in the CLIP-I metric, likely due to significant pose variations in the objects, which may decrease the CLIP-I score. For multi-object generation, our method outperforms other approaches across all evaluation metrics, which we attribute to its ability to effectively avoid object confusion and significantly enhance CLIP-I performance. Furthermore, as shown in Tab.~\ref{tab:dreambench}, we evaluate single-object generation performance on DreamBench. Our method outperforms others in the CLIP-T metric while delivering competitive results in CLIP-I and DINO. We conduct experiments on multi-object cases from CustomConcept101, which we denote as M-CustomConcept101. As shown in Fig.~\ref{tab:customconcept101}, our model achieves the highest performance in the DINO score and ranks second in both the CLIP-T and CLIP-I metrics.

\noindent
\textbf{Qualitative Comparison.} We conduct qualitative comparison experiments on RelationBench to evaluate relation-aware customized image generation. As shown in Fig.~\ref{fig:multi_comparison}, our method outperforms others in generating multi-object relationships, effectively avoiding object confusion while accurately capturing the intended relationships. Additional results can be found in the SM.

\subsection{Ablation study and Analysis}

\noindent
\textbf{Relation-Aware Data Engine.}
Instead of fine-tuning on our curated \((x_k, c_i, c_t)\) pairs. We train a text embedding \( R^* \) using a set of relation-specific data \((x_k, c'_i, c_t)\), where $c'_i$ is cropped from $x_k$. The well-trained text embedding \( R^* \) is injected into the text embedding during inference. The quantitative results in Tab.~\ref{tab:main_ablation} show that our relation-aware data engine outperforms in both CLIP-T and CLIP-R scores.

\noindent
\textbf{Keypoint Matching Loss (KML).} 
We omit KML during fine-tuning. As shown in Fig.~\ref{fig:abl} a), including the KML leads to more accurate and natural relation-aware customized images using the same random seed. Quantitative results, presented in Tab.~\ref{tab:main_ablation}, show a decline across all evaluation metrics when the KML is removed. Additionally, we experimented with different values for $\lambda$ and found that the best performance occurs at $\lambda = 1e-3$.

\noindent
\textbf{Local Token Injection.} 
For local token injection, the visual comparison in Fig.~\ref{fig:abl} b) shows severe object confusion when local tokens are omitted. The quantitative results in Tab.~\ref{tab:main_ablation} demonstrate that local tokens improve relation-aware customization. We conducted an ablation study on the number of local tokens, as presented in Tab.~\ref{tab:num_patches_ablation}, and determined the optimal number to be 16. Additional ablation studies are shown in the SM.
Combining these methods ensures that DreamRelation achieves high-quality relation-aware customized generation.

\begin{table}[t]
    \centering
    \scalebox{0.8}{
        \input{tables/dreambench}
    }  
    \vspace{-5pt}
    \caption{Evaluation Results on DreamBench.}
    \vspace{-10pt}
    \label{tab:dreambench}
\end{table}

\begin{table}[t]
    \centering
    \scalebox{0.8}{
        \input{tables/customconcept101}
    }  
    \vspace{-5pt}
    \caption{Evaluation Results on M-CustomConcept101.}
    \vspace{-10pt}
    \label{tab:customconcept101}
\end{table}

\begin{table}[t]
    \centering
    \scalebox{0.8}{
        \input{tables/ablation_num_patches}
    }  
    \vspace{-5pt}
    \caption{Ablation study on the number of local tokens.}
    \vspace{-10pt}
    \label{tab:num_patches_ablation}
\end{table}

\begin{table}[t]
    \centering
    \scalebox{0.9}{
        \input{tables/ablation_lambda}
    }  
    \vspace{-5pt}
    \caption{Ablation study on $\lambda$, controlling relative weight of KML.}
    \vspace{-10pt}
    \label{tab:lambda_ablation}
\end{table}

%% file: tables/quantitative_comparison.tex
\begin{tabular}{ccccccccc}
 \toprule
 \multirow{2}{*}{Method}& \multicolumn{4}{c}{Single-object} &\multicolumn{4}{c}{Multi-object}  \\
                       & CLIP-T & CLIP-R & CLIP-I & DINO     & CLIP-T & CLIP-R & CLIP-I & DINO \\ \hline
 DreamBooth (SDXL)~\cite{dreambooth}     & \underline{27.8}   & 18.2   & 74.2   & 62.6     & 24.3   & 16.2   & 67.8   & 57.2 \\
 Custom Diffusion~\cite{customdiffusion}      & 26.5   & 15.2   & 73.2   & 58.8     & 20.1   & 15.4   & 64.7   & 55.3 \\
 Cones-V2~\cite{cones2}              & 24.4   & 13.5   & 72.1   & 57.2     & 21.3   & 15.2   & 64.3   & 54.2 \\ \hline
 ELITE~\cite{ELITE}                 & 25.7   & 14.9   & 75.4   & 61.5     & ---    & ---    & ---    & ---  \\
 AnyDoor~\cite{anydoor}               & 24.5   & 14.7   & 77.4   & 62.2     & 21.6   & 14.9   & 69.7   & \underline{59.8} \\
 BLIP-Diffusion~\cite{blip-diffusion}        & 26.2   & 15.7   & 77.4   & 57.7     & ---    & ---    & ---    & ---  \\
 SSR-Encoder~\cite{ssr_encoder}           & 25.5   & 15.9   & \textbf{80.4}   & 59.4     & 24.2   & 14.6   & 72.1   & 56.2 \\
 MS-Diffusion~\cite{ms-diffusion}          & 26.5   & 18.8   & \underline{78.7}   & \textbf{64.5}     & 26.9   & 18.9   & 73.8   & 58.8 \\
 Reversion+MS~\cite{reversion}          & 27.8   & \underline{19.3}   & 77.8   & 63.1     & \underline{27.2}   & \underline{19.2}   & \underline{74.2}   & 59.7 \\
 \hline
 Ours                  & \textbf{30.6}   & \textbf{21.4}   & 77.9   & \underline{63.4}     & \textbf{28.9}   & \textbf{20.4}   & \textbf{75.4}   & \textbf{62.1} \\
\bottomrule
\end{tabular}

%% file: tables/ablation_main.tex
\begin{tabular}{ccccccccc}
 \toprule
 \multirow{2}{*}{Method}    &\multicolumn{4}{c}{Multi-object}  \\
                            & CLIP-T & CLIP-R & CLIP-I & DINO   \\ \hline
 w/o Relation-aware Data    & 27.3   & 19.4   & \underline{75.3}   & 59.8   \\
 w/o Local Token Injection  & \underline{28.5}   & \underline{19.5}   & 75.1   & 59.9   \\
 w/o Keypoint Matching Loss & 27.4   & 19.2   & 75.2   & \underline{61.2}   \\
 \hline
 Full Model                 & \textbf{28.9}   & \textbf{20.4}   & \textbf{75.4}   & \textbf{62.1}   \\
\bottomrule
\end{tabular}

%% file: tables/dreambench.tex
\begin{tabular}{cccc}
 \toprule
 \multirow{2}{*}{Method}& \multicolumn{3}{c}{Single-object} \\
                        & CLIP-T & CLIP-I & DINO  \\ \hline
 DreamBooth (SDXL)~\cite{dreambooth}     & 31.2   & 81.5   & \textbf{69.2}  \\
 Custom Diffusion~\cite{customdiffusion}       & 28.4   & 77.2   & 66.8  \\
 Cones-V2~\cite{cones2}               & 31.0   & 76.5   & 67.2  \\ \hline
 ELITE~\cite{ELITE}                  & 29.8   & 77.4   & 62.5  \\
 AnyDoor~\cite{anydoor}                & 25.5   & \textbf{82.1}   & 67.8  \\
 SSR-Encoder~\cite{ssr_encoder}            & 30.8   & \textbf{82.1}   & 61.2  \\
 MS-Diffusion~\cite{ms-diffusion}           & \underline{31.5}   & 79.3   & 68.2  \\
 \hline
 Ours                   & \textbf{31.6}   & \underline{81.7}   & \underline{68.4}  \\
\bottomrule
\end{tabular}

%% file: tables/customconcept101.tex
\begin{tabular}{cccc}
 \toprule
 \multirow{2}{*}{Method}& \multicolumn{3}{c}{Multi-object} \\
                        & CLIP-T & CLIP-I & DINO  \\ \hline
 DreamBooth (SDXL)~\cite{dreambooth}      & 29.5   & 67.4   & 49.2  \\
 Custom Diffusion~\cite{customdiffusion}       & 28.1   & 66.2   & 48.8  \\
 Cones-V2~\cite{cones2}               & 29.2   & 66.5   & 47.2  \\ \hline
 AnyDoor~\cite{anydoor}                & 20.2   & \textbf{72.1}   & 51.2  \\
 SSR-Encoder~\cite{ssr_encoder}            & \textbf{30.6}   & 71.1   & \underline{52.2}  \\
 $\lambda$-eclipse~\cite{lambda_clipse}      & 29.2   & 68.2   & 48.2  \\
 MS-Diffusion~\cite{ms-diffusion}           & 28.0   & 70.2   & 51.2  \\
 \hline
 Ours                   & \underline{29.7}   & \underline{71.4}   & \textbf{52.3}  \\
\bottomrule
\end{tabular}

%% file: tables/ablation_num_patches.tex

\begin{tabular}{ccccc}
 \toprule
 \multirow{2}{*}{Num Local Tokens} &\multicolumn{4}{c}{Multi-object}\\
                       & CLIP-T & CLIP-R & CLIP-I & DINO \\ \hline
 N=2$\times$2          & \underline{28.3}   & \underline{19.4}   & \underline{75.3}   & \underline{60.1} \\
 N=4$\times$4          & \textbf{28.9}   & \textbf{20.4}   & \textbf{75.4}   & \textbf{62.1} \\ 
 N=8$\times$8          & 27.9   & 18.1   & 74.1   & 59.4 \\
\bottomrule
\end{tabular}

%% file: tables/ablation_lambda.tex

\begin{tabular}{ccccc}
 \toprule
 \multirow{2}{*}{Lambda}&\multicolumn{4}{c}{Multi-object}\\
                        & CLIP-T & CLIP-R & CLIP-I & DINO \\ 
                        \hline
 $\lambda$=1e-2         & \underline{27.5}   & \underline{19.3}   & 72.6   & 59.6 \\
 $\lambda$=1e-3         & \textbf{28.9}   & \textbf{20.4}   & \textbf{75.4}   & \textbf{62.1} \\
 $\lambda$=1e-4         & 26.8   & 18.1   & \underline{73.9}   & \underline{60.4} \\
\bottomrule
\end{tabular}

%% file: contents/5_conclusion.tex
\vspace{-2mm}
\section{Conclusion}
\label{sec:conclusion}
In this work, we introduce a new challenging task: relation-aware customized image generation. 
This task aims to generate objects that adhere to the relationship specified in the text prompt and preserve the identities of the user-provided images. 
To support this, we propose a data engine for generating high-quality fine-tuning data. 
Our method, DreamRelation, incorporates Keypoint Matching Loss and Local Token Injection, effectively capturing relation information and generating natural relations between customized objects. 
We also present a fair comparison with other methods on a new benchmark, RelationBench. 
Extensive experiments demonstrate the effectiveness of our method, highlighting its potential for applications in interactive scenario generation, relation detection, and more.
Our research can inspire the direction of relation-aware generation in the community.

\noindent
\textbf{Acknowledgement.} This work is supported by the National Key Research and Development Program of China (No. 2023YFC3807600).

%% file: contents/6_appendix.tex
\clearpage
\setcounter{page}{1}
\maketitlesupplementary

\noindent
\textbf{Overview.}
The supplementary includes these sections:
\begin{itemize}
    \item \textbf{\cref{sec:supp-introduction-video}.} Introduction video.
    \item \textbf{\cref{sec:supp-experiment-results}.} More experimental results.
    \item \textbf{\cref{sec:supp-implementation-details}.} Implementation details of the experiments.
    \item \textbf{\cref{sec:supp-failure-cases}.} Failure cases and limitations of our model.
    \item \textbf{\cref{sec:supp-relationbench}.} Details of the RelationBench.
    \item \textbf{\cref{sec:supp-i2v}.} Incorporate with CogVideoX-I2V.
    \item \textbf{\cref{sec:supp-social-impact}.} Social impacts.
\end{itemize}

\section{Introduction Video}
\label{sec:supp-introduction-video}
We provide a video introduction of our work. Please refer to ``\textbf{introduction\_video.mp4}'' in the supplementary file.

\section{Additional Experimental Results}
\label{sec:supp-experiment-results}
\begin{figure*}[t]
  \centering
\includegraphics[width=1.0\textwidth]{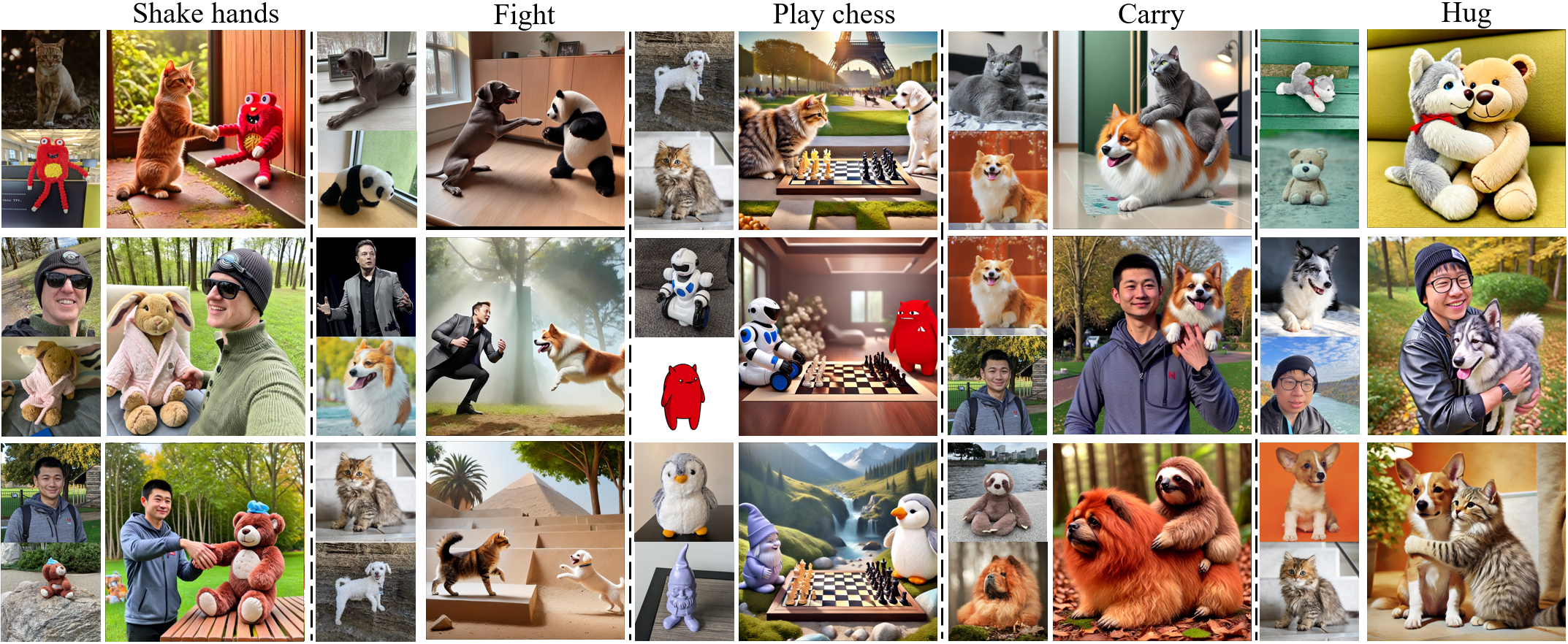}
  \caption{Additional results of relation-aware generation across a wide range of objects.}
  \label{fig:main_results}
\end{figure*}

\begin{figure}[t]
  \centering
  \includegraphics[width=0.45\textwidth]{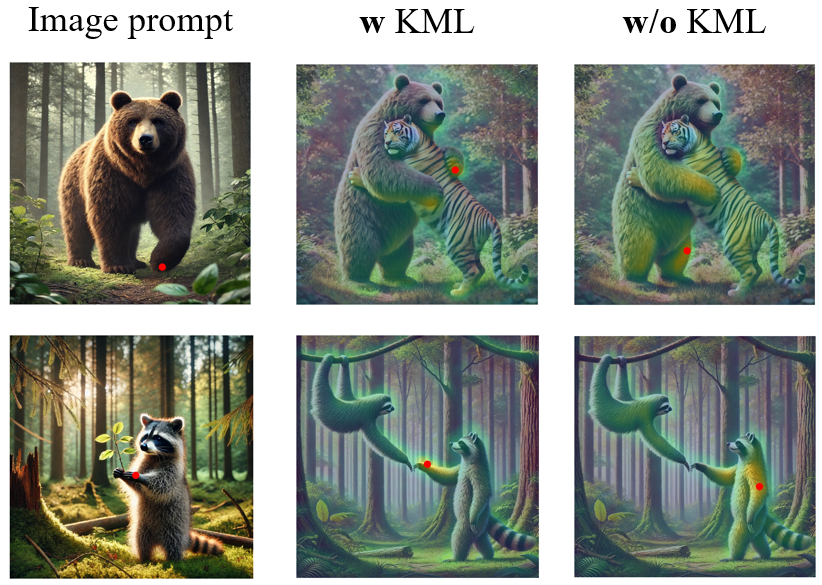}
  \caption{KML enhances the alignment of specific parts of the object in image prompt and model prediction.}
  \label{fig:append_kml}
\end{figure}

\begin{figure}[t]
  \centering
  \includegraphics[width=0.45\textwidth]{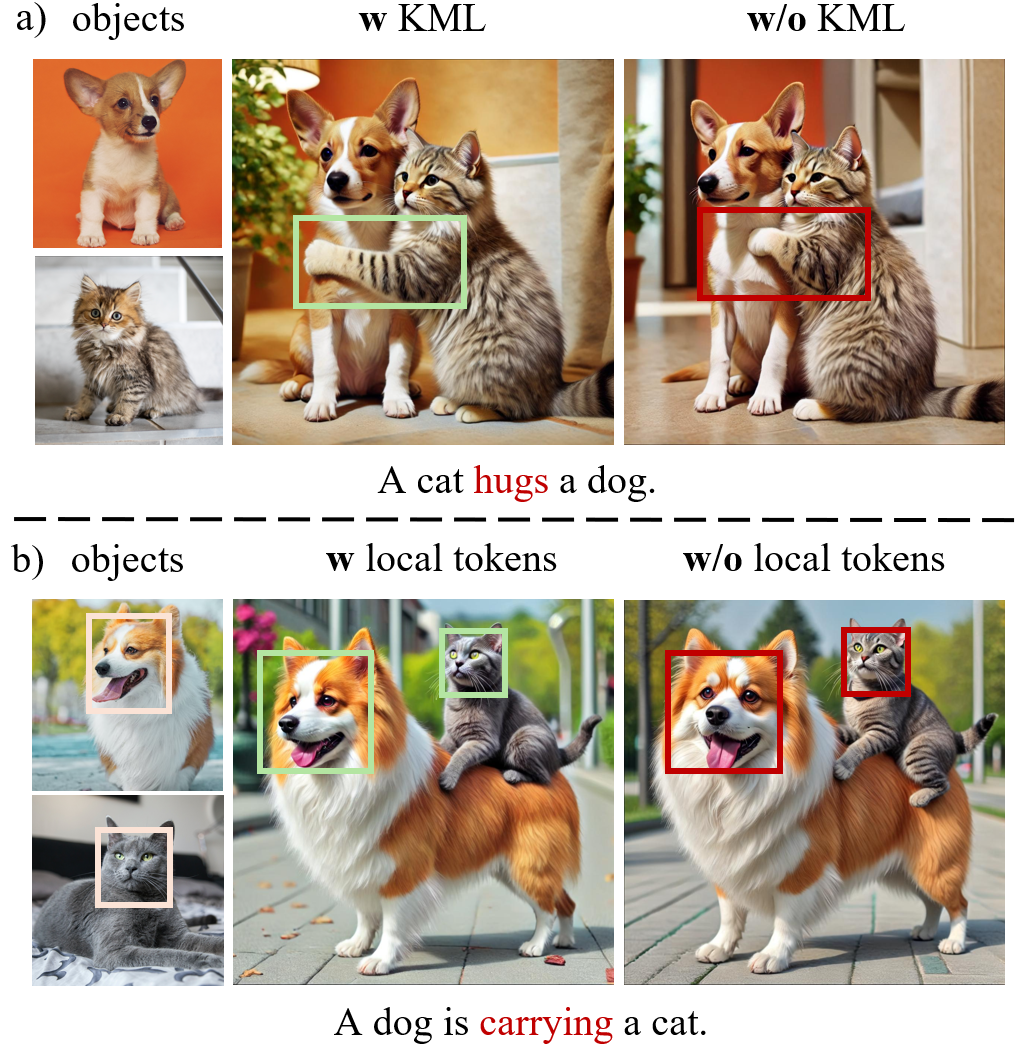}
  \caption{Additional ablation studies on KML and Local Tokens.}
  \label{fig:append_abl}
\end{figure}

\noindent
\textbf{Keypoint Matching Loss.}
We use X-Pose~\cite{unipose} as our keypoint detector due to its open-vocabulary detection capabilities that are compatible with a wide range of objects instead of humans only. The keypoint matching loss (KML) facilitates relation generation by explicitly guiding the model's pose manipulation, resulting in more accurate pose generation. As shown in Fig.~\ref{fig:append_abl}, the cat's arm crosses the dog's body, accurately depicting the "hug" relation. The visualization in Fig.~\ref{fig:append_kml} shows the cosine similarity between the latent representation of the image prompt and model prediction. Implementation with KML shows a better alignment of specific parts during training and inference.

\begin{figure}[t]
  \centering
  \includegraphics[width=0.45\textwidth]{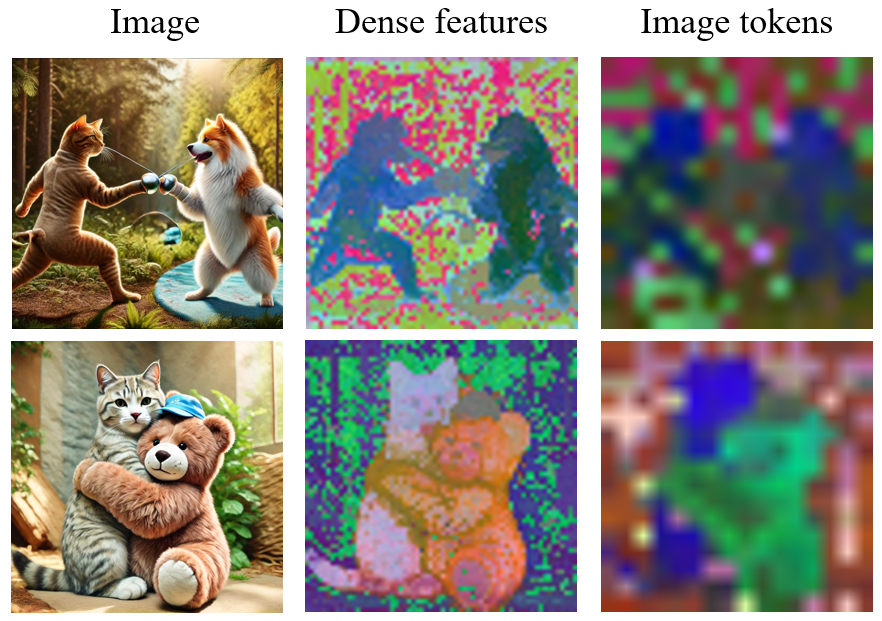}
  \caption{Visualization of Dense feature and Image tokens by Principal Component Analysis (PCA).}
  \label{fig:append_local}
\end{figure}

\noindent
\textbf{Local Tokens Injection.}
To understand why local features enhance relation-aware generation, we adopt Principal Component Analysis (PCA) to project the dense feature more compactly. As shown in Fig.~\ref{fig:append_local}, dense features provide more fine-grained information than CLIP image tokens and provide detailed information about each object to construct the interaction in relation. Aiding in distinguishing between different objects during the generation process and helping to avoid object confusion, especially in cases of heavy overlap. Moreover, it can facilitate object appearance alignment. For the injection method, we note that simply concatenating local tokens with image-level tokens produced the best performance as shown in Fig.~\ref{tab:injection_methods_ablation}.

\begin{figure}[t]
  \centering
\includegraphics[width=0.45\textwidth]{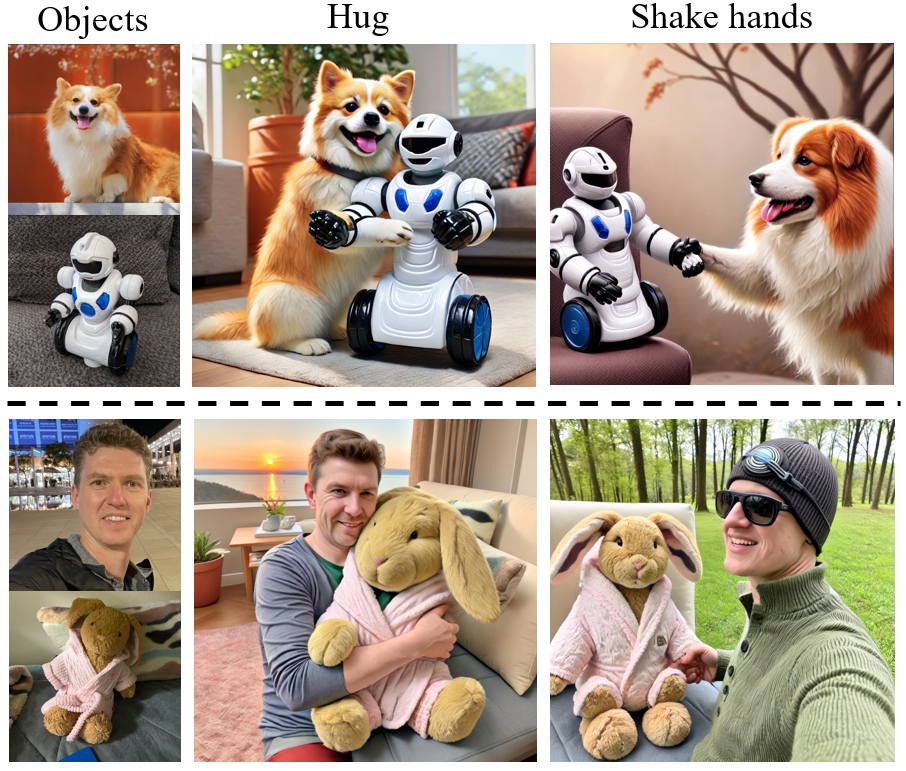}
  \caption{Multi-object relation-aware image customization results of pet, toy, plushie, and person.}
  \label{fig:append_multi1}
\end{figure}

\begin{table}[t]
    \centering
    \caption{Ablation study on Local Image Encoder's architecture.}
    \vspace{-4pt}
    \scalebox{0.8}{\input{tables/ablation_clip_backbone}}  
    \label{tab:clip_backbone_ablation}
\end{table}

\begin{table}[t]
    \centering
    \caption{Ablation study on local token injection methods.}
    \scalebox{0.9}{\input{tables/ablation_injection_method}}  
    \label{tab:injection_methods_ablation}
\end{table}

\noindent
\textbf{Relation-Aware Customization Data Engine.}
Our relation-aware data are generated by DALLE-3~\cite{dalle3}, which can effectively preserve the identity of some categories through its multi-turn dialogue capability. Specifically, we preserve the object's identity by adding "the photo of the same" in the text prompt. Although our relation-aware data only contained animal categories, it can effectively learn the relation information in images and generalize to a wide range of objects as shown in Fig.~\ref{fig:main_results}.

\begin{figure*}[t]
  \centering
\includegraphics[width=1.0\textwidth]{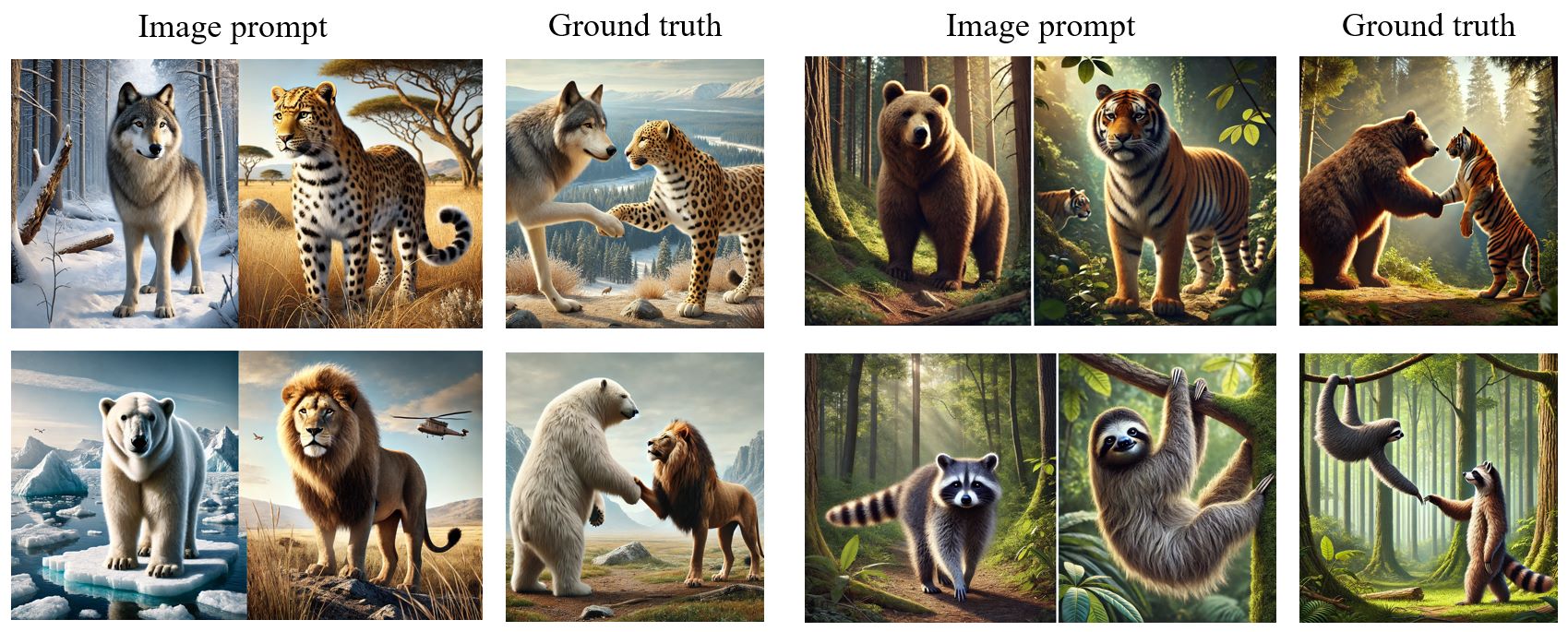}
\vspace{-16pt}
  \caption{Our fine-tuning dataset as an example.}
  \label{fig:append_tuning}
  \vspace{-6pt}
\end{figure*}

\begin{figure*}[t]
  \centering
\includegraphics[width=1.0\textwidth]{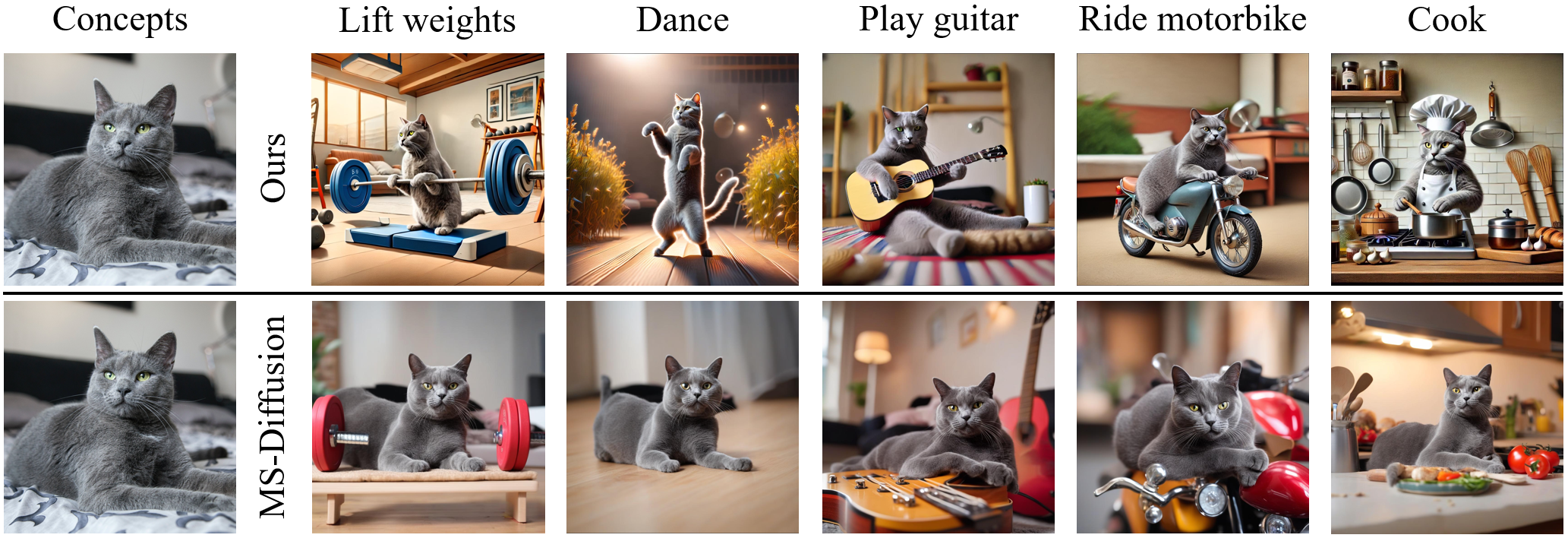}
\vspace{-16pt}
  \caption{Single-object comparison with our base model MS-Diffusion: The results demonstrate that our method generates more accurate and natural relation-aware images.}
  \label{fig:single_comparison}
  \vspace{-6pt}
\end{figure*}

\begin{figure*}[t]
  \centering
\includegraphics[width=1.0\textwidth]{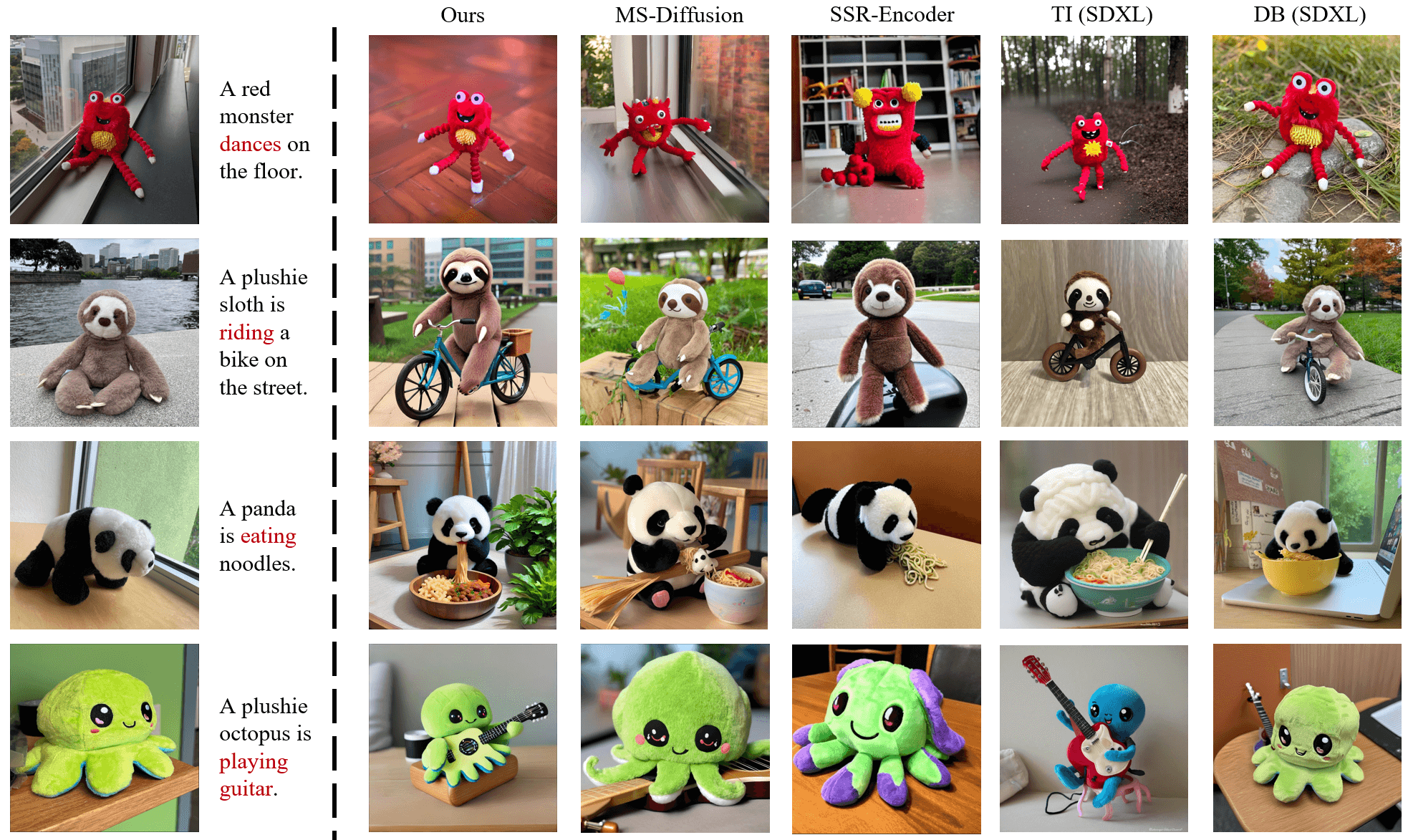}
\vspace{-16pt}
  \caption{Single-object comparison. TI and DB indicate Textual Inversion and DreamBooth, respectively. Our methods achieve the best balance between relation generation and identity preservation.}
  \label{fig:append_single_comparison}
  \vspace{-6pt}
\end{figure*}

\begin{figure*}[t]
  \centering
\includegraphics[width=1.0\textwidth]{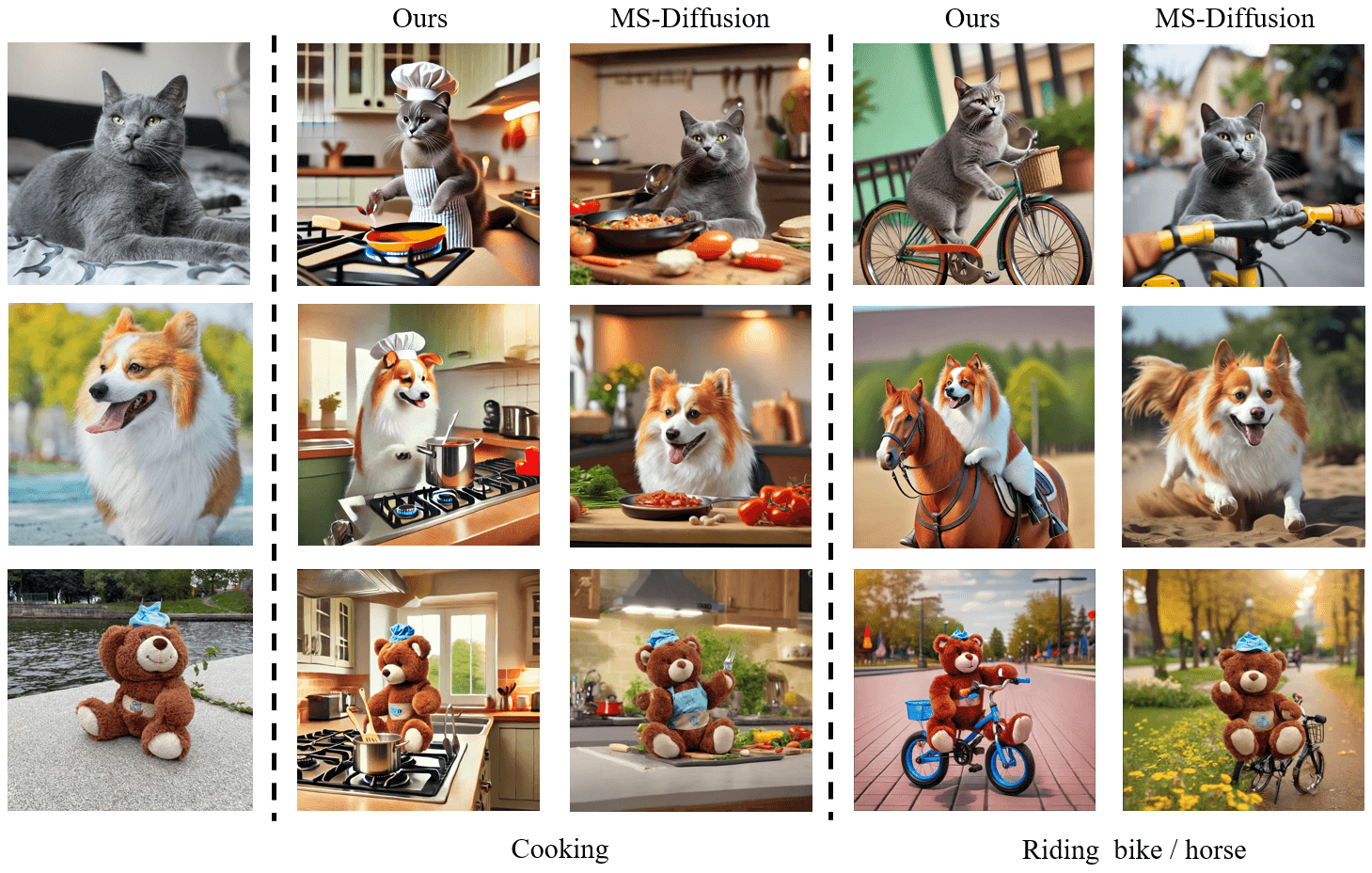}
\vspace{-16pt}
  \caption{Single-object comparison with our base model MS-Diffusion}
  \label{fig:append_single1}
  \vspace{-6pt}
\end{figure*}

\noindent
\textbf{Relation-Aware Customized Image Generation.}
We show more qualitative results in Fig.~\ref{fig:append_single_comparison}, Fig.~\ref{fig:append_single1}, and Fig.~\ref{fig:main_results}. We compare our methods with both training-based and tuning-based methods. Our method shows a clear advantage in pose manipulation to fit the relation. We conduct additional results in Fig.~\ref{fig:single_comparison} to show DreamRelation's differences with our base model-MS-Diffusion.

\begin{figure*}[t]
  \centering
\includegraphics[width=1.0\textwidth]{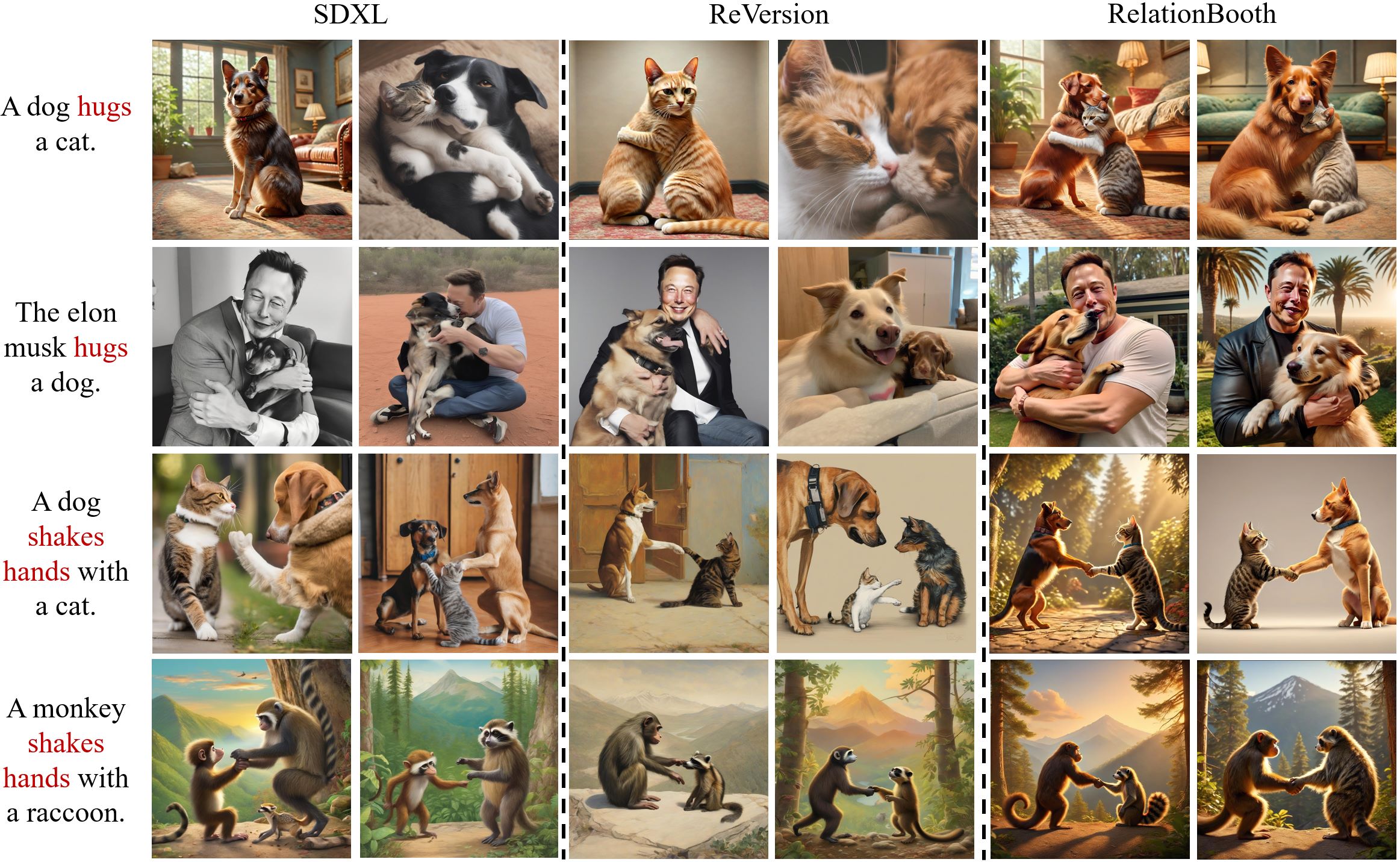}
\vspace{-16pt}
  \caption{Our DreamRelation is compatible with SDXL to address Relation Inversion task.}
  \label{fig:append_relation_inversion}
  \vspace{-6pt}
\end{figure*}

\noindent
\textbf{Relation Inversion Task.}
As shown in Fig.~\ref{fig:append_relation_inversion}, our DreamRelation enhances SDXL's ability to generate images that accurately adhere to specific relations. Compared to ReVersion~\cite{reversion}, our method generates more precise relations, without object confusion or missing, demonstrating the strong performance of DreamRelation in the Relation Generation task.

\noindent
\textbf{Evaluation on DreamBench and CustomConcept101.}
As shown in Tab.~\ref{tab:dreambench} and Tab.~\ref{tab:customconcept101}, we evaluate single-object generation performance on DreamBench. Our method outperforms others in the CLIP-T metric while delivering competitive results in CLIP-I and DINO. To evaluate multi-object generation, we conduct experiments on multi-object cases from CustomConcept101, which we denote as M-CustomConcept101. As shown in Fig.~\ref{tab:customconcept101}, our model achieves the highest performance in the DINO score and ranks second in both the CLIP-T and CLIP-I metrics.

\section{Additional Implementation Details}
\label{sec:supp-implementation-details}
\noindent
\textbf{Local Image Encoder's Implementation Details.}
To enhance region-language alignment of dense features, we employ self-distillation on CLIP-ViT-bigG. The training is conducted on the train2017 split of the COCO dataset for 6 epochs, using 8 A100 GPUs with a batch size of 2 per GPU. We apply the Adam optimizer with a learning rate of $1e-5$ and a weight decay of $0.1$. The Local Image Encoder, containing 1.8B parameters, extracts local tokens that are concatenated with image tokens during fine-tuning and inference to mitigate the confusion problem between objects.

\noindent
\textbf{Baselines' Implementation Details.}
We incorporate ReVersion~\cite{reversion} with MS-Diffusion~\cite{ms-diffusion} by its official implementation. We fine-tuning a learnable text embedding on a set of relation-specific images. We inject the text embedding into text prompt embedding during inference. We implement ReVersion~\cite{reversion} on DreamBooth~\cite{dreambooth} by similar procedure. For tuning-based methods, we implement Textual Inversion~\cite{textual_inversion}, DreamBooth~\cite{dreambooth}, and Custom Diffusion~\cite{customdiffusion} using their respective diffuser versions, with learning rates and tuning steps aligned to those reported in the original papers. We implement Mix-of-Show~\cite{mix-of-show} by their official repository. We utilize the official implementations and checkpoints for training-based methods, adjusting hyperparameters as needed during evaluation. Specifically, we set the scale to 0.6 in MS-Diffusion~\cite{ms-diffusion} and sample 30 steps using the EulerDiscreteScheduler. For the SSR Encoder~\cite{ssr_encoder}, we employ the UniPCMultistepScheduler, sampling 30 steps and adjusting the scale for each object to accommodate different cases. For $\lambda$-Eclipse, we apply the default settings of the official implementation without modification.

\section{Failure cases}
\label{sec:supp-failure-cases}
\begin{figure*}[t]
  \centering
\includegraphics[width=1.0\textwidth]{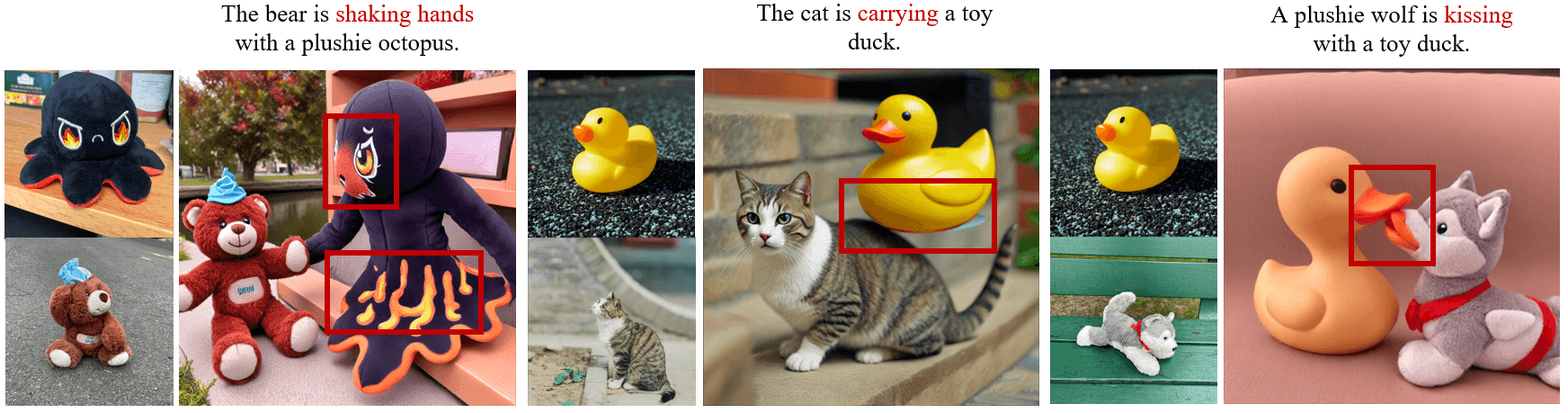}
  \caption{Failure cases of our DreamRelation.}
  \label{fig:append_failure}
\end{figure*}

As shown in Fig.~\ref{fig:append_failure}, we present three typical failure cases from our experimental results. The first involves unreasonable relation generation requests, such as asking a plushie octopus, which lacks limbs, to 'shake hands.' In response, our model generates additional arms for the plushie octopus, leading to a mismatched appearance. The second issue is the unnaturalness of some generated relations, such as a duck failing to make contact with a cat as it should. The final failure case is object confusion at the interaction point, which is a common challenge across all multi-object generation models.

\section{RelationBench}
\label{sec:supp-relationbench}
\begin{figure*}[t]
  \centering
  \includegraphics[width=0.9\textwidth]{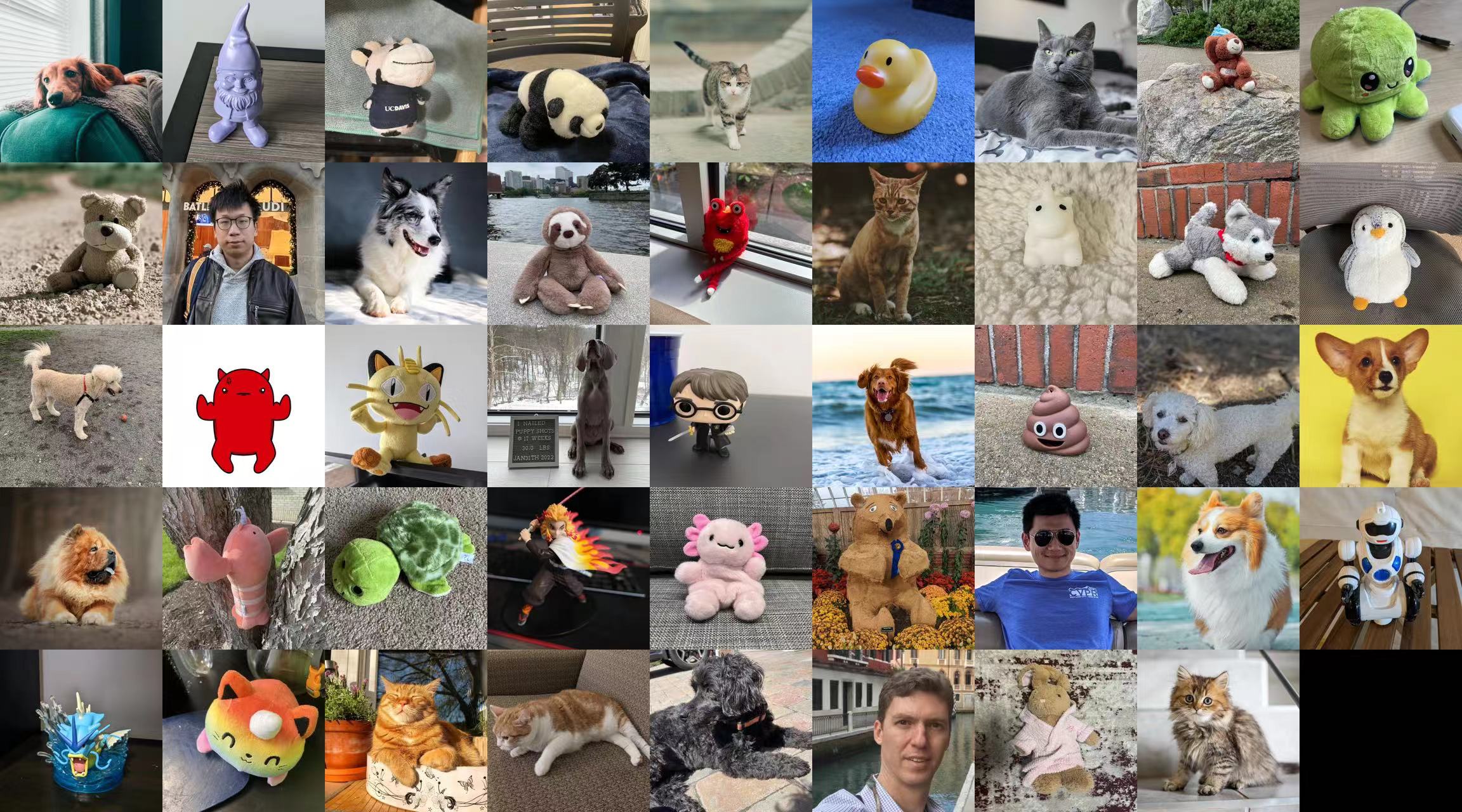}
  \vspace{-10pt}
  \caption{Objects in our proposed RelationBench}
  \label{fig:relationbench}
  \vspace{-4pt}
\end{figure*}

\begin{figure*}[t]
  \centering
  \includegraphics[width=0.5\textwidth]{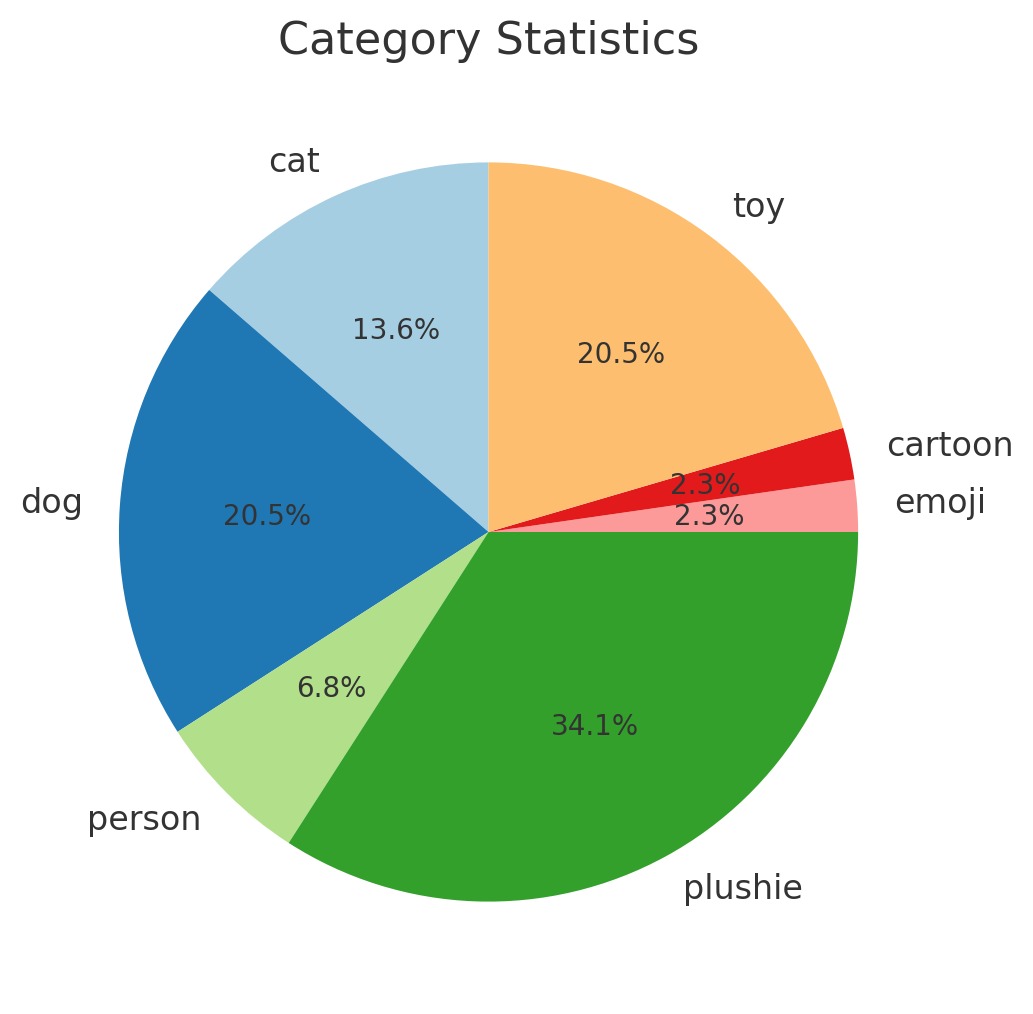}
  \vspace{-16pt}
  \caption{Object category in RelationBench}
  \label{fig:relationbench_category}
  \vspace{-6pt}
\end{figure*}

In this section, we show the objects and text prompts contained in our RelationBench in Fig.~\ref{fig:relationbench}, Fig.~\ref{fig:relationbench_category}, and Tab.~\ref{tab:catpion}. The objects are selected from well-known benchmarks DreamBench and CustomConcept101, covering the commonly seen categories in the real world. The relations in text prompts have covered the most common relations in the real world.

\begin{table*}[t]
\centering
\caption{Text prompt in RelationBench.}
\begin{tabular}{lp{12cm}}
\toprule
\textbf{No.} & \textbf{Prompt} \\ 
\midrule
1 & A \{ \} is playing guitar on a park bench. \\
2 & A \{ \} is playing piano in a grand hall. \\
3 & A \{ \} is eating dinner in a bustling restaurant. \\
4 & A \{ \} is dancing in the moonlight. \\
5 & A \{ \} is lifting weights in a modern gym. \\
6 & A \{ \} is reading a book by the fireplace. \\
7 & A \{ \} is skiing down a steep slope in the Alps, with snowflakes falling gently. \\
8 & A \{ \} is sleeping peacefully in a hammock under the shade of a palm tree. \\
9 & A \{ \} is cooking lunch in a kitchen. \\
10 & A \{ \} is singing on stage during a vibrant music festival. \\
11 & A \{ \} is riding a bike along the scenic countryside road. \\
12 & A \{ \} is riding a horse on the grassland. \\
13 & A \{ \} is riding a motorbike on the street. \\
14 & A \{ \} is playing soccer on football playground. \\
15 & A \{ \} is playing chess with a \{ \} under a tree. \\
16 & A \{ \} is partner dancing with a \{ \} in a vintage ballroom. \\
17 & A \{ \} is carrying a \{ \} on the diving room. \\
18 & A \{ \} is fencing with a \{ \} in an elegant arena. \\
19 & A \{ \} shakes hands with a \{ \} in the forest. \\
20 & A \{ \} is kissing a \{ \}. \\
21 & A \{ \} is playing basketball with a \{ \} on a street court. \\
22 & A \{ \} is wrestling with a \{ \} in a championship ring. \\
23 & A \{ \} is hugging a \{ \} in front of the mountain. \\
24 & A \{ \} is fighting with a \{ \} in a garden. \\
25 & A \{ \} is sitting back to back with a \{ \} on a hilltop. \\
\bottomrule
\label{tab:catpion}
\end{tabular}
\end{table*}

\section{Incorporate with CogVideoX-5b-I2V}
\label{sec:supp-i2v}
\begin{figure*}[t]
  \centering
\includegraphics[width=1.0\textwidth]{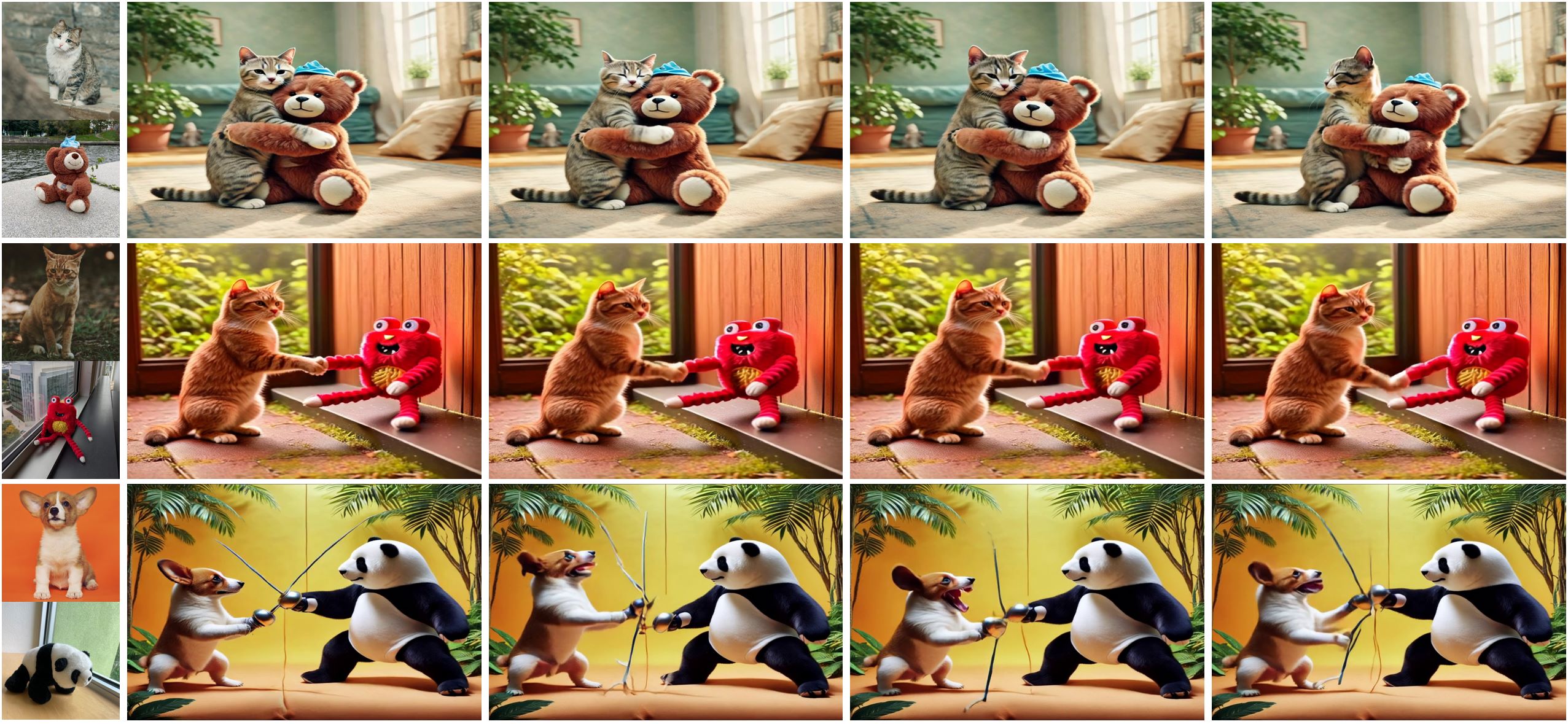}
\vspace{-16pt}
  \caption{Incorporate with CogVideoX-5b-I2V.}
  \label{fig:append_video_frame}
  \vspace{-6pt}
\end{figure*}

Additionally, we use our generated relation-aware customized images as the first frame to generate videos by the CogVideoX-I2V model, the generated results are shown in Fig.~\ref{fig:append_video_frame}. CogVideoX-I2V can handle simple relations such as "shaking hands" but struggles with complex interactions such as "hugging".

\section{Social Impact}
\label{sec:supp-social-impact}
\noindent
\textbf{Positive societal impacts.} 
Relation-aware image customization enables users to generate images that not only contain customized objects but also capture their meaningful relationships. This opens up new opportunities for creative professionals, such as designers, advertisers, and educators, to communicate complex ideas visually with greater precision and flexibility. It has the potential to streamline content creation in diverse fields, from personalized marketing to educational tools, making high-quality, contextually rich imagery accessible without the need for extensive resources.

\noindent
\textbf{Potential negative societal impacts.} 
The ability to generate customized images that involve specific relationships between objects could be misused to fabricate misleading or harmful visual narratives, including false representations of events or manipulative visual content in political or social contexts. Additionally, if the models are trained on biased data, they may reinforce existing societal biases, marginalizing certain groups or distorting the accuracy of represented relationships.

\noindent
\textbf{Mitigation strategies.} 
To reduce misuse, ethical guidelines should be established to govern the responsible development and application of this technology. Promoting transparency about generated content and integrating fairness and diversity considerations into dataset selection are key strategies for mitigating potential harms.

%% file: tables/ablation_clip_backbone.tex

\begin{tabular}{ccccc}
 \toprule
 \multirow{2}{*}{Model}&\multicolumn{4}{c}{Multi-object}\\
                       & CLIP-T & CLIP-R & CLIP-I & DINO \\ \hline
 EVA-CLIP-L14          & \underline{23.6}   & \underline{15.7}   & 56.4   & \underline{54.8} \\
 CLIP-ViT-L14          & 22.9   & 14.8   & \underline{58.3}   & 52.7 \\
 CLIP-ViT-bigG         & \textbf{28.9}   & \textbf{20.4}   & \textbf{75.4}   & \textbf{62.1} \\
\bottomrule
\end{tabular}

%% file: tables/ablation_injection_method.tex

\begin{tabular}{ccccc}
 \toprule
 \multirow{2}{*}{Injection method} &\multicolumn{4}{c}{Multi-object}\\
                       & CLIP-T & CLIP-R & CLIP-I & DINO \\ \hline
 Add                   & 25.4   & \underline{18.5}   & \underline{71.0}   & \underline{56.9} \\
 Linear Projection     & \underline{25.8}   & 18.3   & 68.2   & 54.4 \\
 Concatenate           & \textbf{28.9}   & \textbf{20.4}   & \textbf{75.4}   & \textbf{62.1} \\
\bottomrule
\end{tabular}

%% file: main.bbl
\begin{thebibliography}{45}
\providecommand{\natexlab}[1]{#1}
\providecommand{\url}[1]{\texttt{#1}}
\expandafter\ifx\csname urlstyle\endcsname\relax
  \providecommand{\doi}[1]{doi: #1}\else
  \providecommand{\doi}{doi: \begingroup \urlstyle{rm}\Url}\fi

\bibitem[Bai et~al.(2024)Bai, Ye, Chow, Song, Chen, Li, Dong, Zhu, and Yan]{bai2024meissonic}
Jinbin Bai, Tian Ye, Wei Chow, Enxin Song, Qing-Guo Chen, Xiangtai Li, Zhen Dong, Lei Zhu, and Shuicheng Yan.
\newblock Meissonic: Revitalizing masked generative transformers for efficient high-resolution text-to-image synthesis.
\newblock \emph{arXiv preprint arXiv:2410.08261}, 2024.

\bibitem[Betker et~al.(2023)Betker, Goh, Jing, Brooks, Wang, Li, Ouyang, Zhuang, Lee, Guo, et~al.]{dalle3}
James Betker, Gabriel Goh, Li Jing, Tim Brooks, Jianfeng Wang, Linjie Li, Long Ouyang, Juntang Zhuang, Joyce Lee, Yufei Guo, et~al.
\newblock Improving image generation with better captions.
\newblock \emph{Computer Science.}, 2:\penalty0 3, 2023.

\bibitem[Caron et~al.(2021)Caron, Touvron, Misra, J\'egou, Mairal, Bojanowski, and Joulin]{dino}
Mathilde Caron, Hugo Touvron, Ishan Misra, Herv\'e J\'egou, Julien Mairal, Piotr Bojanowski, and Armand Joulin.
\newblock Emerging properties in self-supervised vision transformers.
\newblock In \emph{Int. Conf. Comput. Vis.}, 2021.

\bibitem[Chen et~al.(2024{\natexlab{a}})Chen, Yu, Ge, Yao, Xie, Wu, Wang, Kwok, Luo, Lu, et~al.]{pixart}
Junsong Chen, Jincheng Yu, Chongjian Ge, Lewei Yao, Enze Xie, Yue Wu, Zhongdao Wang, James Kwok, Ping Luo, Huchuan Lu, et~al.
\newblock Pixart-$\alpha$: Fast training of diffusion transformer for photorealistic text-to-image synthesis.
\newblock In \emph{ICLR}, 2024{\natexlab{a}}.

\bibitem[Chen et~al.(2024{\natexlab{b}})Chen, Huang, Liu, Shen, Zhao, and Zhao]{anydoor}
Xi Chen, Lianghua Huang, Yu Liu, Yujun Shen, Deli Zhao, and Hengshuang Zhao.
\newblock Anydoor: Zero-shot object-level image customization.
\newblock In \emph{CVPR}, 2024{\natexlab{b}}.

\bibitem[Chung et~al.(2022)Chung, Hou, Longpre, Zoph, Tay, Fedus, Li, Wang, Dehghani, Brahma, et~al.]{t5}
Hyung~Won Chung, Le Hou, Shayne Longpre, Barret Zoph, Yi Tay, William Fedus, Yunxuan Li, Xuezhi Wang, Mostafa Dehghani, Siddhartha Brahma, et~al.
\newblock Scaling instruction-finetuned language models.
\newblock \emph{arXiv preprint arXiv:2210.11416}, 2022.

\bibitem[Gal et~al.(2023)Gal, Alaluf, Atzmon, Patashnik, Bermano, Chechik, and Cohen{-}Or]{textual_inversion}
Rinon Gal, Yuval Alaluf, Yuval Atzmon, Or Patashnik, Amit~Haim Bermano, Gal Chechik, and Daniel Cohen{-}Or.
\newblock An image is worth one word: Personalizing text-to-image generation using textual inversion.
\newblock In \emph{ICLR}, 2023.

\bibitem[Gu et~al.(2024)Gu, Wang, Wu, Shi, Chen, Fan, Xiao, Zhao, Chang, Wu, et~al.]{mix-of-show}
Yuchao Gu, Xintao Wang, Jay~Zhangjie Wu, Yujun Shi, Yunpeng Chen, Zihan Fan, Wuyou Xiao, Rui Zhao, Shuning Chang, Weijia Wu, et~al.
\newblock Mix-of-show: Decentralized low-rank adaptation for multi-concept customization of diffusion models.
\newblock \emph{NeurIPS}, 2024.

\bibitem[Hu et~al.(2022)Hu, Shen, abd Zeyuan Allen-Zhu, Li, Wang, Wang, and Chen]{lora}
Edward~J Hu, Yelong Shen, Phillip~Wallis abd Zeyuan Allen-Zhu, Yuanzhi Li, Shean Wang, Lu Wang, and Weizhu Chen.
\newblock Lora: Low-rank adaptation of large language models.
\newblock In \emph{ICLR}, 2022.

\bibitem[Huang et~al.(2024)Huang, Gong, Feng, Chen, Fu, Liu, and Wang]{adi}
Siteng Huang, Biao Gong, Yutong Feng, Xi Chen, Yuqian Fu, Yu Liu, and Donglin Wang.
\newblock Learning disentangled identifiers for action-customized text-to-image generation.
\newblock In \emph{CVPR}, 2024.

\bibitem[Huang et~al.(2023)Huang, Wu, Jiang, Chan, and Liu]{reversion}
Ziqi Huang, Tianxing Wu, Yuming Jiang, Kelvin~CK Chan, and Ziwei Liu.
\newblock Reversion: Diffusion-based relation inversion from images.
\newblock \emph{arXiv preprint arXiv:2303.13495}, 2023.

\bibitem[Kingma and Ba(2014)]{adam}
Diederik~P Kingma and Jimmy Ba.
\newblock Adam: A method for stochastic optimization.
\newblock \emph{arXiv preprint arXiv:1412.6980}, 2014.

\bibitem[Kirillov et~al.(2023)Kirillov, Mintun, Ravi, Mao, Rolland, Gustafson, Xiao, Whitehead, Berg, Lo, Doll{\'{a}}r, and Girshick]{kirillov2023sam}
Alexander Kirillov, Eric Mintun, Nikhila Ravi, Hanzi Mao, Chlo{\'{e}} Rolland, Laura Gustafson, Tete Xiao, Spencer Whitehead, Alexander~C. Berg, Wan{-}Yen Lo, Piotr Doll{\'{a}}r, and Ross~B. Girshick.
\newblock Segment anything.
\newblock In \emph{ICCV}, 2023.

\bibitem[Kong et~al.(2024)Kong, Wu, Hu, Han, Peng, Xu, Luo, Zhang, Wang, and Fu]{anymaker}
Lingjie Kong, Kai Wu, Xiaobin Hu, Wenhui Han, Jinlong Peng, Chengming Xu, Donghao Luo, Jiangning Zhang, Chengjie Wang, and Yanwei Fu.
\newblock Anymaker: Zero-shot general object customization via decoupled dual-level id injection.
\newblock \emph{arXiv preprint arXiv:2406.11643}, 2024.

\bibitem[Kumari et~al.(2023)Kumari, Zhang, Zhang, Shechtman, and Zhu]{customdiffusion}
Nupur Kumari, Bingliang Zhang, Richard Zhang, Eli Shechtman, and Jun{-}Yan Zhu.
\newblock Multi-concept customization of text-to-image diffusion.
\newblock In \emph{CVPR}, 2023.

\bibitem[Li et~al.(2023)Li, Li, and Hoi]{blip-diffusion}
Dongxu Li, Junnan Li, and Steven C.~H. Hoi.
\newblock Blip-diffusion: Pre-trained subject representation for controllable text-to-image generation and editing.
\newblock In \emph{NeurIPS}, 2023.

\bibitem[Lian et~al.(2023)Lian, Li, Yala, and Darrell]{llm_grounded_diffusion}
Long Lian, Boyi Li, Adam Yala, and Trevor Darrell.
\newblock Llm-grounded diffusion: Enhancing prompt understanding of text-to-image diffusion models with large language models.
\newblock \emph{arXiv preprint arXiv:2305.13655}, 2023.

\bibitem[Lin et~al.(2024)Lin, Chen, Shi, Zhu, Liang, Miao, Jin, Zhao, Wu, Yan, et~al.]{clif}
Wang Lin, Jingyuan Chen, Jiaxin Shi, Yichen Zhu, Chen Liang, Junzhong Miao, Tao Jin, Zhou Zhao, Fei Wu, Shuicheng Yan, et~al.
\newblock Non-confusing generation of customized concepts in diffusion models.
\newblock \emph{ICML}, 2024.

\bibitem[Liu et~al.(2024)Liu, Li, Wu, and Lee]{llava}
Haotian Liu, Chunyuan Li, Qingyang Wu, and Yong~Jae Lee.
\newblock Visual instruction tuning.
\newblock \emph{NeurIPS}, 2024.

\bibitem[Liu et~al.(2023)Liu, Zhang, Shen, Zheng, Zhu, Feng, Liu, Zhao, Zhou, and Cao]{cones2}
Zhiheng Liu, Yifei Zhang, Yujun Shen, Kecheng Zheng, Kai Zhu, Ruili Feng, Yu Liu, Deli Zhao, Jingren Zhou, and Yang Cao.
\newblock Cones 2: Customizable image synthesis with multiple subjects.
\newblock In \emph{NeurIPS}, 2023.

\bibitem[Materzynska et~al.(2023)Materzynska, Sivic, Shechtman, Torralba, Zhang, and Russell]{customizing_motion}
Joanna Materzynska, Josef Sivic, Eli Shechtman, Antonio Torralba, Richard Zhang, and Bryan Russell.
\newblock Customizing motion in text-to-video diffusion models.
\newblock \emph{arXiv preprint arXiv:2312.04966}, 2023.

\bibitem[mengmeng Ge et~al.(2024)mengmeng Ge, Jia, Isobe, Li, Wang, Mu, Zhou, liwang Amd, Lu, Tian, Sirasao, and Barsoum]{customizing_with_inverted_interaction}
mengmeng Ge, Xu Jia, Takashi Isobe, Xiaomin Li, Qinghe Wang, Jing Mu, Dong Zhou, liwang Amd, Huchuan Lu, Lu Tian, Ashish Sirasao, and Emad Barsoum.
\newblock Customizing text-to-image generation with inverted interaction.
\newblock In \emph{ACM MM}, 2024.

\bibitem[Pang et~al.(2024)Pang, Yin, Zhao, Wu, Wang, Li, and Mao]{attndreambooth}
Lianyu Pang, Jian Yin, Baoquan Zhao, Feize Wu, Fu~Lee Wang, Qing Li, and Xudong Mao.
\newblock Attndreambooth: Towards text-aligned personalized text-to-image generation.
\newblock \emph{arXiv preprint arXiv:2406.05000}, 2024.

\bibitem[Patel et~al.(2024)Patel, Jung, Baral, and Yang]{lambda_clipse}
Maitreya Patel, Sangmin Jung, Chitta Baral, and Yezhou Yang.
\newblock $\lambda$-eclipse: Multi-concept personalized text-to-image diffusion models by leveraging clip latent space.
\newblock \emph{arXiv preprint arXiv:2402.05195}, 2024.

\bibitem[Peebles and Xie(2023)]{DiT}
William Peebles and Saining Xie.
\newblock Scalable diffusion models with transformers.
\newblock In \emph{ICCV}, 2023.

\bibitem[Pernias et~al.(2023)Pernias, Rampas, Richter, Pal, and Aubreville]{StableCascade}
Pablo Pernias, Dominic Rampas, Mats~Leon Richter, Christopher Pal, and Marc Aubreville.
\newblock W{\"u}rstchen: An efficient architecture for large-scale text-to-image diffusion models.
\newblock In \emph{ICLR}, 2023.

\bibitem[Podell et~al.(2023)Podell, English, Lacey, Blattmann, Dockhorn, M{\"u}ller, Penna, and Rombach]{sdxl}
Dustin Podell, Zion English, Kyle Lacey, Andreas Blattmann, Tim Dockhorn, Jonas M{\"u}ller, Joe Penna, and Robin Rombach.
\newblock Sdxl: Improving latent diffusion models for high-resolution image synthesis.
\newblock \emph{arXiv preprint arXiv:2307.01952}, 2023.

\bibitem[Radford et~al.(2021)Radford, Kim, Hallacy, Ramesh, Goh, Agarwal, Sastry, Askell, Mishkin, Clark, Krueger, and Sutskever]{clip}
Alec Radford, Jong~Wook Kim, Chris Hallacy, Aditya Ramesh, Gabriel Goh, Sandhini Agarwal, Girish Sastry, Amanda Askell, Pamela Mishkin, Jack Clark, Gretchen Krueger, and Ilya Sutskever.
\newblock Learning transferable visual models from natural language supervision.
\newblock In \emph{ICML}, 2021.

\bibitem[Ren et~al.(2024)Ren, Li, Chen, Pei, Shao, Guo, Peng, Song, and Zhu]{ultrapixel}
Jingjing Ren, Wenbo Li, Haoyu Chen, Renjing Pei, Bin Shao, Yong Guo, Long Peng, Fenglong Song, and Lei Zhu.
\newblock Ultrapixel: Advancing ultra-high-resolution image synthesis to new peaks.
\newblock \emph{arXiv preprint arXiv:2407.02158}, 2024.

\bibitem[Rombach et~al.(2022)Rombach, Blattmann, Lorenz, Esser, and Ommer]{stable_diffusion}
Robin Rombach, Andreas Blattmann, Dominik Lorenz, Patrick Esser, and Bj{\"{o}}rn Ommer.
\newblock High-resolution image synthesis with latent diffusion models.
\newblock In \emph{CVPR}, 2022.

\bibitem[Ruiz et~al.(2023)Ruiz, Li, Jampani, Pritch, Rubinstein, and Aberman]{dreambooth}
Nataniel Ruiz, Yuanzhen Li, Varun Jampani, Yael Pritch, Michael Rubinstein, and Kfir Aberman.
\newblock Dreambooth: Fine tuning text-to-image diffusion models for subject-driven generation.
\newblock In \emph{CVPR}, 2023.

\bibitem[Saharia et~al.(2022)Saharia, Chan, Saxena, Li, Whang, Denton, Ghasemipour, Lopes, Ayan, Salimans, Ho, Fleet, and Norouzi]{imagen}
Chitwan Saharia, William Chan, Saurabh Saxena, Lala Li, Jay Whang, Emily~L. Denton, Seyed Kamyar~Seyed Ghasemipour, Raphael~Gontijo Lopes, Burcu~Karagol Ayan, Tim Salimans, Jonathan Ho, David~J. Fleet, and Mohammad Norouzi.
\newblock Photorealistic text-to-image diffusion models with deep language understanding.
\newblock In \emph{NeurIPS}, 2022.

\bibitem[Vaswani et~al.(2017)Vaswani, Shazeer, Parmar, Uszkoreit, Jones, Gomez, Kaiser, and Polosukhin]{attention}
Ashish Vaswani, Noam Shazeer, Niki Parmar, Jakob Uszkoreit, Llion Jones, Aidan~N Gomez, {\L}ukasz Kaiser, and Illia Polosukhin.
\newblock Attention is all you need.
\newblock In \emph{NeurIPS}, 2017.

\bibitem[Voynov et~al.(2023)Voynov, Chu, Cohen-Or, and Aberman]{p+}
Andrey Voynov, Qinghao Chu, Daniel Cohen-Or, and Kfir Aberman.
\newblock p+: Extended textual conditioning in text-to-image generation.
\newblock \emph{arXiv preprint arXiv:2303.09522}, 2023.

\bibitem[Wang et~al.(2024)Wang, Fu, Huang, He, and Jiang]{ms-diffusion}
X Wang, Siming Fu, Qihan Huang, Wanggui He, and Hao Jiang.
\newblock Ms-diffusion: Multi-subject zero-shot image personalization with layout guidance.
\newblock \emph{arXiv preprint arXiv:2406.07209}, 2024.

\bibitem[Wei et~al.(2023)Wei, Zhang, Ji, Bai, Zhang, and Zuo]{ELITE}
Yuxiang Wei, Yabo Zhang, Zhilong Ji, Jinfeng Bai, Lei Zhang, and Wangmeng Zuo.
\newblock {ELITE:} encoding visual concepts into textual embeddings for customized text-to-image generation.
\newblock In \emph{ICCV}, 2023.

\bibitem[Wei et~al.(2024)Wei, Zhang, Qing, Yuan, Liu, Liu, Zhang, Zhou, and Shan]{dreamvideo}
Yujie Wei, Shiwei Zhang, Zhiwu Qing, Hangjie Yuan, Zhiheng Liu, Yu Liu, Yingya Zhang, Jingren Zhou, and Hongming Shan.
\newblock Dreamvideo: Composing your dream videos with customized subject and motion.
\newblock In \emph{CVPR}, 2024.

\bibitem[Wu et~al.(2024{\natexlab{a}})Wu, Li, Zeng, Zhang, Zhou, Li, Tong, and Chen]{wu2024motionbooth}
Jianzong Wu, Xiangtai Li, Yanhong Zeng, Jiangning Zhang, Qianyu Zhou, Yining Li, Yunhai Tong, and Kai Chen.
\newblock Motionbooth: Motion-aware customized text-to-video generation.
\newblock \emph{arXiv preprint arXiv:2406.17758}, 2024{\natexlab{a}}.

\bibitem[Wu et~al.(2024{\natexlab{b}})Wu, Zhang, Xu, Jin, Li, Liu, and Loy]{clipself}
Size Wu, Wenwei Zhang, Lumin Xu, Sheng Jin, Xiangtai Li, Wentao Liu, and Chen~Change Loy.
\newblock Clipself: Vision transformer distills itself for open-vocabulary dense prediction.
\newblock \emph{ICLR}, 2024{\natexlab{b}}.

\bibitem[Yang et~al.(2023)Yang, Gu, Zhang, Zhang, Chen, Sun, Chen, and Wen]{paint_by_example}
Binxin Yang, Shuyang Gu, Bo Zhang, Ting Zhang, Xuejin Chen, Xiaoyan Sun, Dong Chen, and Fang Wen.
\newblock Paint by example: Exemplar-based image editing with diffusion models.
\newblock In \emph{CVPR}, 2023.

\bibitem[Yang et~al.(2024)Yang, Zeng, Zhang, and Zhang]{unipose}
Jie Yang, Ailing Zeng, Ruimao Zhang, and Lei Zhang.
\newblock Unipose: Detecting any keypoints.
\newblock \emph{ECCV}, 2024.

\bibitem[Ye et~al.(2023)Ye, Zhang, Liu, Han, and Yang]{ip-adapter}
Hu Ye, Jun Zhang, Sibo Liu, Xiao Han, and Wei Yang.
\newblock Ip-adapter: Text compatible image prompt adapter for text-to-image diffusion models.
\newblock \emph{arXiv preprint arxiv:2308.06721}, 2023.

\bibitem[Zhang et~al.(2024)Zhang, Song, Liu, Wang, Yu, Tang, Li, Tang, Hu, Pan, et~al.]{ssr_encoder}
Yuxuan Zhang, Yiren Song, Jiaming Liu, Rui Wang, Jinpeng Yu, Hao Tang, Huaxia Li, Xu Tang, Yao Hu, Han Pan, et~al.
\newblock Ssr-encoder: Encoding selective subject representation for subject-driven generation.
\newblock In \emph{CVPR}, 2024.

\bibitem[Zhou et~al.(2022)Zhou, Loy, and Dai]{dense_clip}
Chong Zhou, Chen~Change Loy, and Bo Dai.
\newblock Extract free dense labels from clip.
\newblock In \emph{ECCV}, 2022.

\bibitem[Zhu et~al.(2024)Zhu, Li, Ma, He, and Xiu]{multibooth}
Chenyang Zhu, Kai Li, Yue Ma, Chunming He, and Li Xiu.
\newblock Multibooth: Towards generating all your concepts in an image from text.
\newblock \emph{arXiv preprint arXiv:2404.14239}, 2024.

\end{thebibliography}
